\documentclass[10pt,twocolumn,letterpaper]{article}

\usepackage{cvpr}              

\usepackage{graphicx}
\usepackage{amsmath}
\usepackage{amssymb}
\usepackage{booktabs}

\usepackage{times}
\usepackage{epsfig}
\usepackage{graphicx}
\usepackage{algorithm}
\usepackage{listings}
\usepackage[british,american]{babel}
\usepackage{enumitem}
\usepackage{multirow}
\usepackage[table]{xcolor}
\usepackage{ctable} 

\definecolor{mygray}{gray}{0.9}
\definecolor{mygray2}{gray}{0.7}

\newcommand{\app}{\raise.17ex\hbox{$\scriptstyle\sim$}}

\usepackage{tabulary}
\newcolumntype{x}[1]{>{\centering\arraybackslash}p{#1pt}}

\definecolor{green}{HTML}{39b54a}  
\definecolor{red}{HTML}{ea4335}  
\newcommand{\hlg}[1]{\textcolor{green}{#1}}
\newcommand{\hlr}[1]{\textcolor{red}{#1}}

\newcommand{\better}[2]{{#1} {\fontsize{7pt}{1em}\selectfont \hlg{$\uparrow$#2}}}
\newcommand{\betterdown}[2]{{#1} {\fontsize{7pt}{1em}\selectfont \hlg{$\downarrow$#2}}}

\newcommand{\worseup}[2]{{#1} {\fontsize{7pt}{1em}\selectfont \hlr{$\uparrow$#2}}}

\usepackage[pagebackref,breaklinks,colorlinks]{hyperref}

\usepackage[capitalize]{cleveref}
\crefname{section}{Sec.}{Secs.}
\Crefname{section}{Section}{Sections}
\Crefname{table}{Table}{Tables}
\crefname{table}{Tab.}{Tabs.}

\begin{document}

\title{LEMaRT: Label-Efficient Masked Region Transform for Image Harmonization}

\author{Sheng Liu \quad Cong Phuoc Huynh \quad Cong Chen \quad Maxim Arap \quad Raffay Hamid\\
Amazon Prime Video\\
{\tt\small \{shenlu, conghuyn, checongt, maxarap, raffay\}@amazon.com}
}
\maketitle

\begin{abstract}
\noindent We present a simple yet effective self-supervised pre-training method for image harmonization which can leverage large-scale unannotated image datasets. To achieve this goal, we first generate pre-training data online with our \textbf{L}abel-\textbf{E}fficient \textbf{Ma}sked \textbf{R}egion \textbf{T}ransform (LEMaRT) pipeline. Given an image, LEMaRT generates a foreground mask and then applies a set of transformations to perturb various visual attributes, \eg, defocus blur, contrast, saturation, of the region specified by the generated mask. We then pre-train image harmonization models by  recovering the original image from the perturbed image. Secondly, we introduce an image harmonization model, namely SwinIH, by retrofitting the Swin Transformer~\cite{Liu2021Swin} with a combination of local and global self-attention mechanisms. Pre-training SwinIH with LEMaRT results in a new state of the art for image harmonization, while being label-efficient, \ie, consuming less annotated data for fine-tuning than existing methods. Notably, on iHarmony4 dataset~\cite{Cong2020DoveNet}, SwinIH outperforms the state of the art, \ie, SCS-Co~\cite{Hang2022SCSCo} by a margin of $0.4$ dB when it is fine-tuned on only $50\%$ of the training data, and by $1.0$ dB when it is trained on the full training dataset.
\end{abstract}

\section{Introduction}
\label{sec:intro}

\begin{figure}
\centering
\includegraphics[width=1.0\linewidth]{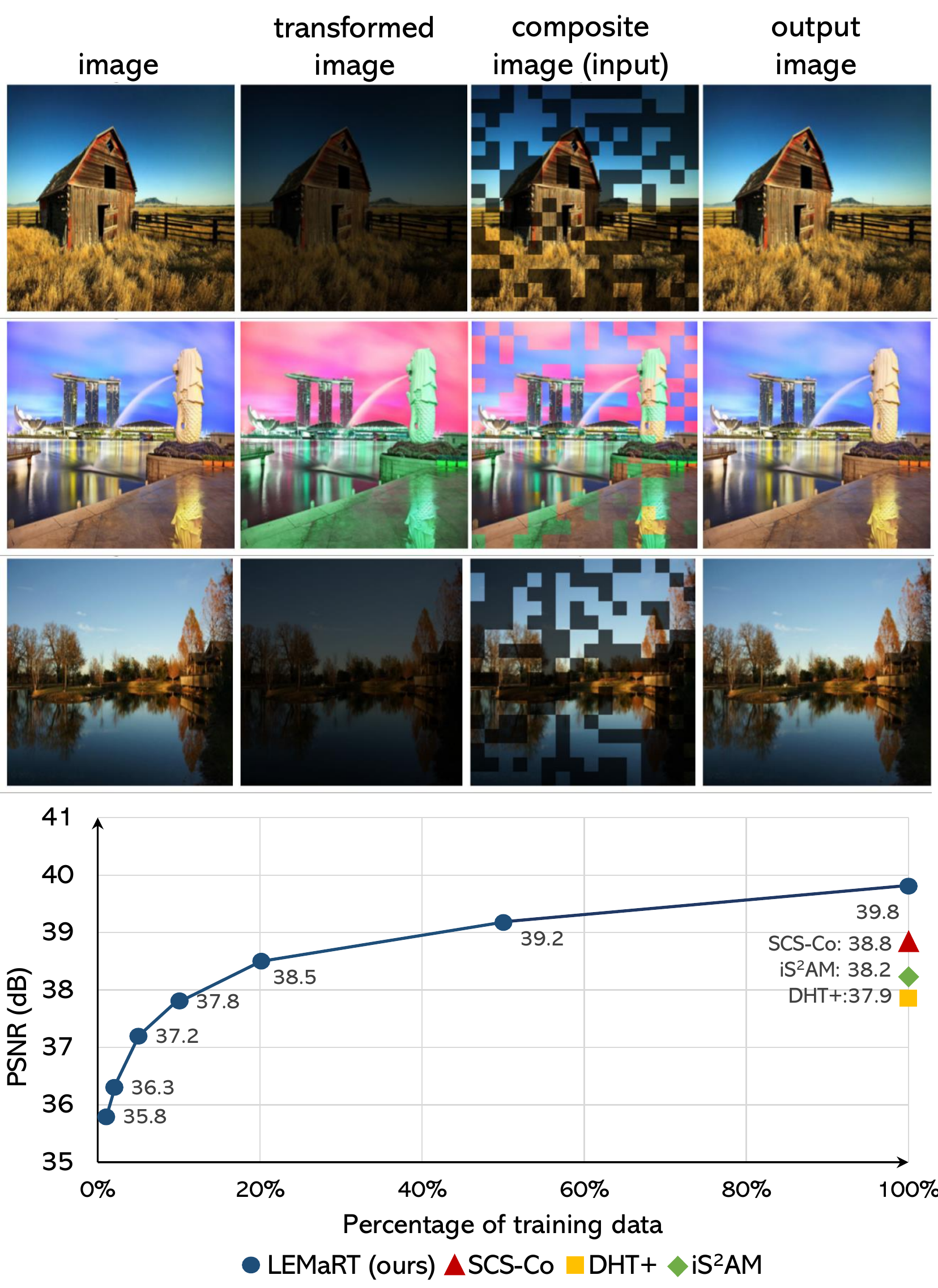}
\caption{Top: given an image, LEMaRT applies a set of transformations, \eg, brightness, hue adjustment, to obtain a transformed image. The transformed image is then combined with the original image to form a composite image, which is used to pre-train our SwinIH image harmonization model. As shown in the right-hand column, SwinIH is capable of reconstructing photo-realistic output images after pre-training and fine-tuning. Bottom: using our \textit{LEMaRT} pre-training scheme, our image harmonization model (\textit{SwinIH}) surpasses state of the art (SOTA) counterparts with less than $40$\% of the training data from iHarmony4 for fine-tuning.}
\vspace{-14pt}
\label{fig:data_efficiency}
\end{figure}

\noindent The goal of image harmonization is to synthesize photo-realistic images by extracting and transferring foreground regions from an image to another (background) image. The main challenge is the appearance mismatch between the foreground and the surrounding background, due to differences in camera and lens settings, capturing conditions, such as illumination, and post-capture image processing. Image harmonization aims to resolve this mismatch by adjusting the appearance of the foreground in a composite image to make it compatible with the background. Research in image harmonization has relevant applications in photo-realistic image editing and enhancement~\cite{Zhou2019DPR,Zhu2020Indomain}, video synthesis~\cite{Lee2019VidIntoVid,Wang2018Vid2Vid} and data augmentation for various computer vision tasks~\cite{Dwibedi2017CutPasteLearn,Wang2019PSIS,Ghiasi2021SimpleCopyPaste}.

Traditional image harmonization approaches perform color transforms to match the low-level color statistics of the foreground to the background with the aim to achieve photorealism~\cite{Xue2012Understanding,Lalonde2007ColorCompatibility,Sunkavalli2010MultiscaleIH,Song2020Illumination}. However, the generalization ability of these methods is questionable because the evaluation was only conducted at a small scale, mainly using human judgement. More recent works~\cite{Cong2020DoveNet} have constructed real image harmonization datasets with tens to thousands of images to train learning-based methods. However, due to the bottleneck of manual editing, these datasets do not match the scale often required to train large-scale neural networks. Rendered image  datasets~\cite{Cao2022DIH,Guo2021Intrinsic} are more scalable but they suffer from the domain gap between synthetic and real images. As a result, the performance of image harmonization models is constrained by the limited size of a few existing datasets~\cite{Jiang2021SSH,Cong2020DoveNet} on which they can be trained.

Inspired by the impressive performance leap achieved by pre-trained models~\cite{Radford2013CLIP,He2021MAE} on various downstream tasks, \eg, image classification, object detection, image captioning, in this work, we introduce a novel self-supervised pre-training method to boost the performance of image harmonization models while being label-efficient, \ie, consuming small amounts of fine-tuning data. The novelty of our technique lies in the use of foreground masking strategies and the perturbation of foreground visual attributes to self-generate training data without annotations. Hence, we name our pre-training method as \textbf{L}abel-\textbf{E}fficient \textbf{Ma}sked \textbf{R}egion \textbf{T}ransform (LEMaRT). In the first step, LEMaRT proposes pseudo foreground regions in an image. Subsequently, it applies a set of transformations to perturb visual attributes of the foreground, including contrast, sharpness, blur and  saturation. These transformations aim to mimic the appearance discrepancy between the foreground and the background. Using the transformed image, \ie, image with the perturbed foreground, as the input, LEMaRT pre-trains image harmonization models to reconstruct the original image, as shown in the top half of Figure~\ref{fig:data_efficiency}.

Subsequently, we design an image harmonization model based on Swin Transformer~\cite{Liu2021Swin}, namely SwinIH, which is short for Swin Image Harmonization. We build our model upon Swin Transformer instead of the ViT model~\cite{dosovitskiy2021an} mainly due to the efficiency gain offered by its local shifted window (Swin) attention. Similar to the design of the original Swin Transformer, we keep the local self-attention mechanism in all the Transformer blocks up except the last one, where we employ global self-attention. We introduce global self-attention into SwinIH to alleviate block boundary artifacts produced by the Swin Transformer model when it is directly trained for image harmonization.

We verify that LEMaRT consistently improves the performance of models with a range of vision Transformer and CNN architectures compared to training only on the target dataset, \eg, iHarmony4. When we pre-train our SwinIH model on MS-COCO dataset with LEMaRT and then fine-tune it on iHarmony4~\cite{Cong2020DoveNet}, it outperforms the state of the art~\cite{Hang2022SCSCo} by $0.4$ dB while using only $\textit{50}\%$ of the samples from iHarmony4 for fine-tuning, and by $1.0$ dB when using all the samples (see the plot in the bottom half of Figure~\ref{fig:data_efficiency}).

\vspace{0.1cm} \noindent The \textbf{key contributions} of our work are summarized below.
\setlength\itemsep{0.05em}

\noindent $\bullet$ We introduce Label-Efficient Masked Region Transform (LEMaRT), a novel pre-training method for image harmonization, which is able to leverage large-scale unannotated image datasets. 

\noindent $\bullet$ We design SwinIH, an image harmonization model based on the Swin Transformer architecture~\cite{Liu2021Swin}.

\noindent $\bullet$ LEMaRT (SwinIH) establishes new state of the art on iHarmony4 dataset, while consuming significantly less amount of training data. LEMaRT also boosts the performance of models with various network architectures.

\begin{figure*}[t]
\centering
\includegraphics[width=0.78\linewidth]{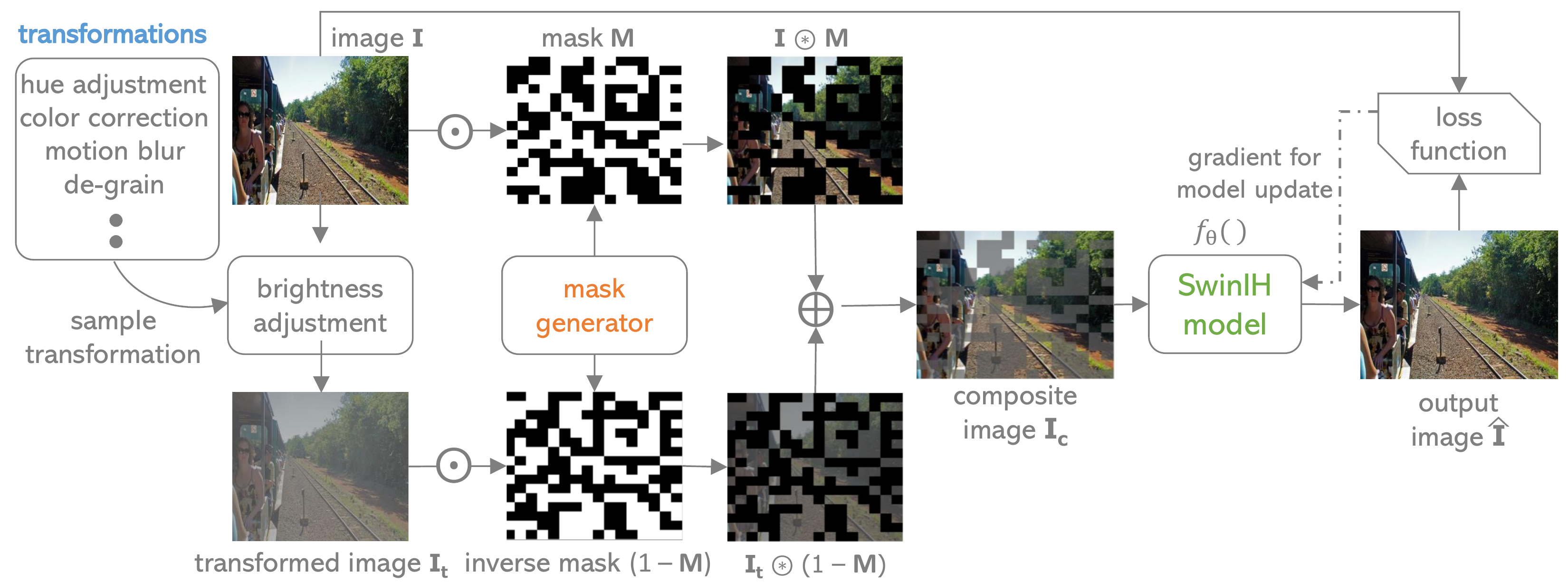}
\caption{Our online data generation and pre-training pipeline (LEMaRT). LEMaRT generates the input composite image $\mathbf{I}_c$ for the pre-training process via masked region transform. The goal of pre-training is to learn an image harmonization model $f_{\mathbf{\theta}}(\cdot)$, \textit{\eg} our SwinIH model, that can reconstruct the original image $\mathbf{I}$ from the composite image $\mathbf{I}_c$.
}
\vspace{-10pt}
\label{fig:tae}
\end{figure*}

\section{Related Work}

\vspace{1pt}
\noindent \textbf{a. Image Harmonization:} Most early works extract and match low-level color statistics of the foreground and its surrounding background. These works rely on color histograms~\cite{Xue2012Understanding}, multi-level pyramid representations~\cite{Sunkavalli2010MultiscaleIH}, color clusters~\cite{Lalonde2007ColorCompatibility}, etc. The limited representation power of low-level features negatively affects their performance.

More recent works~\cite{Jiang2021SSH,Cong2020DoveNet} have constructed datasets at a reasonable scale to advance learning-based methods. Numerous supervised deep learning-based image harmonization models have been trained on these datasets~\cite{cun2020improving,Ling2021RAINNet,Guo2021Intrinsic,Guo_2021_ICCV}. Tsai~\etal~\cite{tsai2017deep} combine image harmonization and semantic segmentation under a multi-task setting. S$^2$AM~\cite{cun2020improving} proposes to predict a foreground mask and to adjust the appearance of foreground with spatial-separated attention. RainNet~\cite{Ling2021RAINNet} transfers statistics of instance normalization layers from the background to the foreground. In addition, generative models have also been trained for image harmonization~\cite{Cong2020DoveNet,Chen2019Toward,Zhu2015Learning}. 

Some state of the art (SOTA) methods formulate image harmonization as a style transfer problem. These methods learn a domain representation of the foreground and background with contrastive learning ~\cite{cong2021bargainnet} or by maximizing mutual information between the foreground and background~\cite{Liang2022RegionCL}. More recently, Hang~\etal~\cite{Hang2022SCSCo} have advanced state of the art results by adding background and foreground style consistency constraints and dynamically sampling negative examples within a contrastive learning paradigm. Using only a reconstruction loss during pre-training and fine-tuning, our method is able to outperform ~\cite{Hang2022SCSCo} with a much simpler training set up.

\vspace{3pt}
\noindent \textbf{b. Transfer Learning:} Transfer learning is a well-known and effective technique for adapting a pre-trained model to a downstream task, especially with limited training data \cite{he2020momentum,chen2020simple,chen2021empirical}. Recent advances in foundation models ~\cite{Radford2013CLIP,Jia2021ALIGN,wang2022omnivl,zhou2021uc2,yuan2021florence,zhong2022regionclip,alayrac2022flamingo} have resulted in models that can be adapted to a wide range of downstream tasks. Sofiiuk~\etal~\cite{sofiiuk2021foreground} propose an image harmonization model which takes visual features extracted from a pre-trained segmentation model as an auxiliary input. Instead of leveraging a pre-trained segmentation model for feature extraction, we specifically pre-train a model for image harmonization. We opt for this direction based on the hypothesis that pre-training for the same target task results in better performance than pre-training for a different task. Inspired by~\cite{he2020momentum}, our LEMaRT method is more suitable for image harmonization than \cite{he2020momentum} because LEMaRT creates training samples by applying transformations to the foreground rather than masking the foreground, which makes the pre-training task closer to image harmonization. In addition, \cite{he2020momentum} introduces an asymmetric encoder-decoder architecture, while our SwinIH model is specifically designed for image harmonization and does not have an explicit encoder or a decoder.

\section{Method}

\subsection{Problem Formulation}
\noindent The goal of image harmonization is to synthesize photo-realistic images by extracting and transferring foreground regions from an image $\mathbf{I}_1$, specified by a binary mask $\mathbf{M}$, to another (background) image $\mathbf{I}_2$. Let $\mathbf{I}_{c} = \mathbf{M} \odot \mathbf{I}_1 \oplus (1 - \mathbf{M}) \odot \mathbf{I}_2$ be the composite image generated by a direct copy and paste of the foreground region from $\mathbf{I}_1$ on top of $\mathbf{I}_2$. The operators $\odot$ and $\oplus$ denote element-wise multiplication and addition, respectively. Subsequently, an image harmonization function $f(\cdot)$ transforms the composite image $\mathbf{I}_{c}$ into a harmonized image $\hat{\mathbf{I}} = f(\mathbf{I}_{c})$, such that the latter is photo-realistic. Deep learning-based image harmonization methods implement this function as a neural network $f_\theta(\cdot)$ with parameters denoted by  $\mathbf{\theta}$. Our goal is to learn $\mathbf{\theta}$ via self-supervised pre-training, so that the function $f_{\mathbf{\theta}}(\cdot)$ can generate photo-realistic  images.

\begin{figure*}[tbh]
\centering
\includegraphics[width=0.98\linewidth]{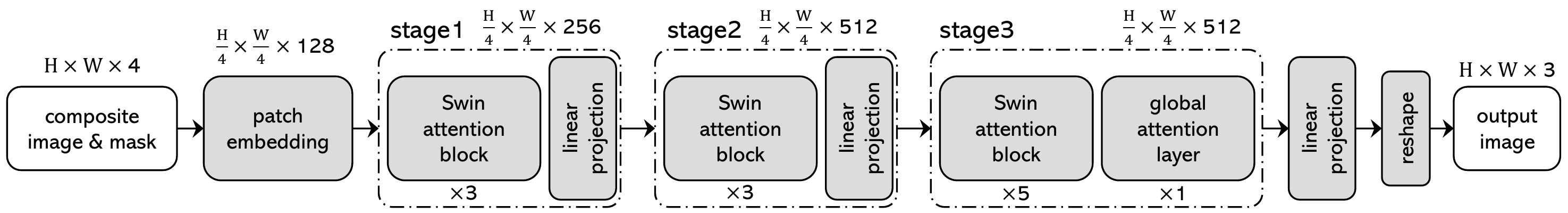}
\vspace{-1pt}
\caption{Illustration of our SwinIH model, \textit{i.e.}, a Transformer-based image harmonization model.}
\vspace{-6pt}
\label{fig:transformer}
\end{figure*}

\subsection{Online Pre-training Data Generation}
\label{sec:mrt}

\noindent We first introduce our data generation and pre-training pipeline, \ie, LEMaRT that generates the input and the ground truth for the pre-training process without relying on any manual annotations. As shown in Figure \ref{fig:tae}, LEMaRT applies a set of random transformations such as hue, contrast, brightness adjustment and defocus blur to perturb the original image $\mathbf{I}$. The generated image is referred to as the transformed image $\mathbf{I}_t$. The random transformations are designed to mimic different kinds of visual mismatches between a foreground region and a background image. In addition, LEMaRT employs a mask generation strategy to propose a foreground mask $\mathbf{M}$ (please refer to $\S$~\ref{sec:mask} for more details). The mask and the set of transformations are generated on the fly for each input image $\mathbf{I}$. With these ingredients, we generate a composite image $\mathbf{I}_c$ from the original image $\mathbf{I}$ and the transformed image $\mathbf{I}_t$ as $\mathbf{I}_{c} = \mathbf{M} \odot \mathbf{I}_t \oplus (1 - \mathbf{M}) \odot \mathbf{I}$. We note that the composite image $\mathbf{I}_c$ is generated and fed into the network in an \textit{online} fashion. 

The goal of pre-training is to learn a harmonization function $f_{\mathbf{\theta}}(\cdot)$ to resolve the mismatch of visual appearance between the foreground and the background of the composite image $\mathbf{I}_{c}$. We formulate this task as the reconstruction of the original image $\mathbf{I}$ from the composite image $\mathbf{I}_{c}$. The original image $\mathbf{I}$ serves as the supervision signal for the pre-training process under the assumption that visual elements of real images are in harmony. This formulation is applicable to a wide range of network architectures, \textit{e.g.}, SwinIH (see $\S$~\ref{sec:model} for details), ViT~\cite{dosovitskiy2021an} and CNN models~\cite{Lee2019VidIntoVid,SunXLW19}.

\subsection{Mask Generation}
\label{sec:mask}

\begin{figure}
    \centering
    \includegraphics[width=0.98\linewidth]{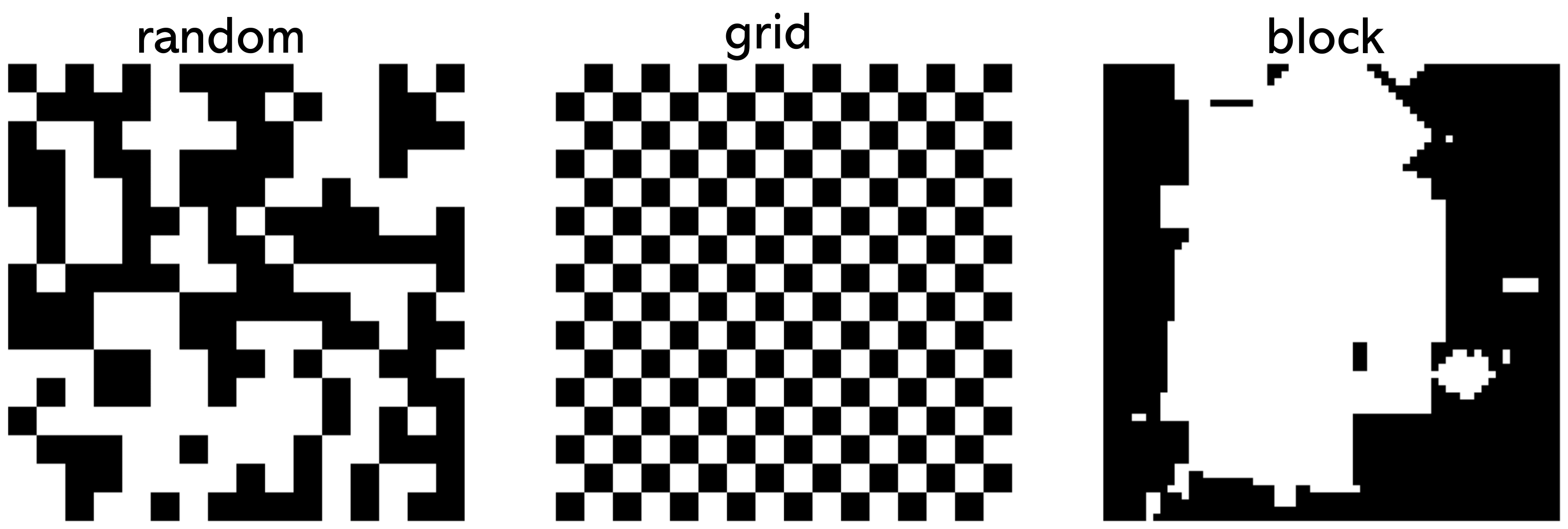}
    \vspace{-1pt}
    \caption{Sample masks generated by the three strategies, \textit{i.e.}, random, grid and block, introduced in $\S$~\ref{sec:mask}.}
    \vspace{-8pt}
    \label{fig:mask}
\end{figure}

\noindent 
We now present three foreground mask generation strategies, which we refer to as \textit{random}, \textit{grid} and \textit{block}. 

\noindent $\bullet$ \textit{random}: as shown in Figure \ref{fig:mask}, this strategy first partitions an image into a regular pattern that consists of $m \times m$ even patches (labelled by white pixels). It then generates a mask by randomly selecting a subset of the image patches.

\noindent $\bullet$ \textit{grid}: similar to \textit{random}, this strategy first partitions an image into regular pattern of $m \times m$ even patches. It then generates a \textit{fixed} mask (same for all images) by selecting image patches following the pattern shown in Figure~\ref{fig:mask}.

\noindent $\bullet$ \textit{block}: inspired by \cite{Bao2021BEiT}, we design a mask generation strategy that attempts to mimic the \textit{shape} of objects. It generates a mask in an iterative manner. Each iteration has two steps, \ie, (1) generating a rectangular region; (2) applying a random homography to the rectangular region to make the boundary of the region be composed of slanted lines. If the rectangular region is smaller than the size of the desired region to be masked, we generate a new region by executing the two steps once more and merge the newly generated region with previously generated regions (please refer to the supplementary materials for more details). 

\subsection{Network Architecture}
\label{sec:model}
\noindent Since the pre-training process is agnostic to network architecture, the only constraint for model design is that it needs to generate an output image of the same size as the input. To this end, we choose to implement our image harmonization model based on a Transformer architecture, due to the recent successes of vision transformer models in various tasks, including image harmonization~\cite{Guo_2021_ICCV,guo2022transformer}.

We choose to build our SwinIH model upon Swin attention blocks~\cite{Liu2021Swin} due to its improved performance and efficiency over global attention layers used by ViT-style models~\cite{dosovitskiy2021an,Guo_2021_ICCV,guo2022transformer}. Let $H$ and $W$ denote the height and width of an input image $\mathbf{I}_{c}$ and $N$ denote the size of an image patch. The length of the sequence of visual tokens is $\frac{H \cdot W}{N^{2}}$. A global attention layer has a space and time complexity of $\mathcal{O}(\frac{H^{2} \cdot W^{2}}{N^{2}})$. On the contrary, the space and time complexity of a Swin attention block is $\mathcal{O}(\frac{H \cdot W \cdot K^{2}}{N^{2}})$, where the shifted window size $K$ is smaller than $H$ and $W$. 

The architecture of our SwinIH model is shown in Figure ~\ref{fig:transformer}. It takes a four-channel image (a channel-wise concatenation of a composite image $\mathbf{I}_c$ and a foreground mask $\mathbf{M}$) as input and generates an output image $\hat{\mathbf{I}}$. Our SwinIH model is composed of three stages. The first two stages consist of three Swin attention blocks. The third stage has five Swin attention blocks and a global attention layer. We set the dimension of the patch embedding to $128$ and double it at the end of the first two stages with linear projections. 

Unlike Swin Transformer\cite{Liu2021Swin} which processes its input in a multi-scale manner\footnote{Swin Transformer\cite{Liu2021Swin} uses a Patch Merging layer to reduce the spatial size of its input by a factor of two at the end of each stage.}, we choose to preserve the original resolution of the input. As will be shown in $\S$~ \ref{sec:exp-network}, such a design choice is important as the information loss due to the reduced resolution hurts model accuracy.

\subsection{Objective Function}
\noindent We adopt the mean squared error (MSE) between the network's output $\hat{\mathbf{I}}$ and the original image $\mathbf{I}$ as the objective function for pre-training. When we fine-tune the pre-trained network, we follow \cite{sofiiuk2021foreground} and use a foreground-normalized MSE loss as the objective function. 

\setlength{\tabcolsep}{3pt}
\renewcommand{\arraystretch}{1.1}
\begin{table*}[tbh]
	\centering
	\begin{tabular}{c|c|ccccccccccc}
        \specialrule{.1em}{.05em}{.05em} 
		\rowcolor{mygray}  &  & composite & DIH & S$^2$AM & DoveNet & BargNet & IntrHarm  & RainNet & iS$^2$AM & DHT+ & SCS-Co & LEMaRT \\
		\rowcolor{mygray} dataset & metric & image &~\cite{tsai2017deep} &~\cite{cun2020improving} &~\cite{Cong2020DoveNet} &~\cite{cong2021bargainnet} &~\cite{Guo2021Intrinsic} &~\cite{Ling2021RAINNet} &~\cite{sofiiuk2021foreground} &~\cite{guo2022transformer} &~\cite{Hang2022SCSCo} & (SwinIH) \\
        \hline \hline
		\multirow{2}[2]{*}{HCOCO} & PSNR$\uparrow$  & 33.9 & 34.7 & 35.5 & 35.8 & 37.0 & 37.2 & 37.1 & {39.2} & 39.2 & \textit{39.9} & \better{\textbf{41.0}}{1.1}  \\
		& MSE$\downarrow$   & 69.4 & 51.9 & 41.1 & 36.7 & 24.8 & 24.9 & 29.5 & {16.5} & 15.0 & \textit{13.6} & \betterdown{\textbf{10.1}}{3.5} \\
        \hline\hline
		\multirow{2}[2]{*}{HAdobe} & PSNR$\uparrow$  & 28.2 & 32.3 & 33.8 & 34.3 & 35.3 & 35.2  & 36.2 & {38.1} & 37.2 & \textit{38.3} & \better{\textbf{39.4}}{1.1} \\
		& MSE$\downarrow$   & 345.5 & 92.7 & 63.4  & 52.3 & 39.9 & 43.0 & 43.4 & {21.9} & 36.8 & \textit{21.0} & \betterdown{\textbf{18.8}}{2.2} \\
        \hline\hline
		\multirow{2}[2]{*}{HFlickr} & PSNR$\uparrow$  & 28.3 & 29.6 & 30.0 & 30.2 & 31.3 & 31.3 & 31.6 & {33.6} & 33.6 & \textit{34.2} & \better{\textbf{35.3}}{1.1} \\
		& MSE$\downarrow$   & 264.4 & 163.4 & 143.5 & 133.1 & 97.3 & 105.1 & 110.6 & {69.7} & 67.9 & \textit{55.8} & \betterdown{\textbf{40.7}}{15.1} \\
        \hline\hline
		\multirow{2}[2]{*}{HD2N} & PSNR$\uparrow$  & 34.0 & 34.6 & 34.5  & 35.3 & 35.7 & 36.0 & 34.8 & {37.7} & 36.4 & \textit{37.8} & \better{\textbf{38.1}}{0.3} \\
		& MSE$\downarrow$   & 109.7 & 82.3 & 76.6 & 52.0 & 51.0 & 55.5 & 57.4 & \textbf{40.6} & 49.7 & \textit{41.8} & \worseup{42.3}{1.7} \\
        \hline\hline
		\multirow{2}[2]{*}{all} & PSNR  & 31.6 & 33.4 & 34.3 & 34.8 & 35.9 & 35.9 & 36.1 & {38.2} & 37.9 & \textit{38.8} & \better{\textbf{39.8}}{1.0} \\
		& MSE$\downarrow$   & 172.5 & 76.8 & 59.7 & 52.3 & 37.8 & 38.7 & 40.3 & {24.4} & 27.9 & \textit{21.3} & \betterdown{\textbf{16.8}}{4.5} \\
        \specialrule{.1em}{.05em}{.05em} 
	\end{tabular}
    \vspace{-4pt}
	\caption[]{Our pre-trained image harmonization model, LEMaRT, outperforms state-of-the-art (SOTA) models on iHarmony4. The column named \textit{composite image} shows the result for the direct copy and paste of foreground regions on top of background images.}
	\label{tab:sota-ihd}%
	\vspace{-6pt}
\end{table*}%

\section{Experiments}
\noindent We evaluate our method by comparing its performance with other state-of-the-art (SOTA) methods and provide insights into our method through ablation studies. We adopt four metrics, \ie, mean squared error (MSE), peak signal to noise ratio (PSNR), foreground mean squared error (fMSE~)\cite{Cong2020DoveNet}, and foreground peak signal to noise ratio (fPSNR)~\cite{Cong2020DoveNet}.

\subsection{Datasets}
\noindent Following previous works~\cite{Cong2020DoveNet, cong2021bargainnet, Guo2021Intrinsic}, we evaluate our method on iHarmony4 dataset~\cite{Cong2020DoveNet}. For completeness, we also evaluate our method on RealHM dataset~\cite{Jiang2021SSH}. Unless otherwise stated, we pre-train our LEMaRT model on the set of 120K unlabeled images from the MS COCO dataset~\cite{Lin2014} and fine-tune on iHarmony4. There is no overlap between the images used for pre-training and the images used for fine-tuning and evaluation. Images in iHarmony4 either come from the set of labeled images in MS COCO, which is disjoint from the unlabeled images in MS COCO, or from other datasets. Following \cite{tsai2017deep,Cong2020DoveNet,Guo2021Intrinsic,cong2021bargainnet,sofiiuk2021foreground}, we resize the input images and the ground truth images to $256 \times 256$.

\subsection{Implementation Details}
\noindent We use an AdamW optimizer~\cite{loshchilov2018fixing} both during pre-training and fine-tuning. We set $\beta_{1}=0.9$, $\beta_{2}=0.95$, $\epsilon=1e^{-8}$ and weight decay to $0.05$. The window size and the patch size of SwinIH are set to $32$ and $4$, respectively. We pre-train our model for $30$ epochs with a batch size of $192$ and a learning rate of $2.7e^{-2}$. We then fine-tune the pre-trained model for $120$ epochs with a learning rate of $2.7e^{-3}$. A cosine annealing scheduler controls the change of learning rate. The minimum learning rate is set to $0.0$. We adopt the random mask generation strategy and set mask ratio to $50$\% during pre-training. This is the default setting for the experiments.

\subsection{Comparison with SOTA Methods}

\vspace{4pt}
\noindent \textbf{a. On iHarmony4 Dataset}
\vspace{1pt}

\noindent In Table~\ref{tab:sota-ihd}, we present a comparison between the performance of our method, LEMaRT (SwinIH), and the performance of existing methods on iHarmony4. Overall, LEMaRT comprehensively outperforms existing methods across the two metrics (PSNR and MSE). Most notably, our method achieves a PSNR of $39.8$ dB, which is $1.0$ dB higher than the previous best method. The MSE of our method is $16.8$, which is $4.5$ lower ($21.1$\% relative improvement) than the previous best method~\cite{Hang2022SCSCo}~\footnote{Our method also outperforms SOTA methods across the iHarmony4 dataset in terms of fPSNR and fMSE. For brevity, we omit them in Table~\ref{tab:sota-ihd} and include them in supplementary materials instead.}.

We notice that our method, LEMaRT, consistently achieves better performance than SOTA methods~\cite{Hang2022SCSCo,guo2022transformer,sofiiuk2021foreground,Ling2021RAINNet,Guo2021Intrinsic,cong2021bargainnet,cun2020improving,tsai2017deep,Cong2020DoveNet} on three of the four subsets, \ie, HCOCO, HAdobe and HFlickr of iHarmony4. Meanwhile, on the HD2N subset, the performance of our method is on par with SOTA methods. While our method yields higher PSNR, the MSE of our method is higher. We hypothesize that the domain of MS COCO, the dataset which we use to pre-train LEMaRT, is not closely aligned with that of HD2N. For example, mountains and buildings are the salient objects in most images in the HD2N subset. However, they do not often appear as the main objects in MS COCO images. 

\begin{figure*}
    \centering
    \includegraphics[width=0.985\linewidth]{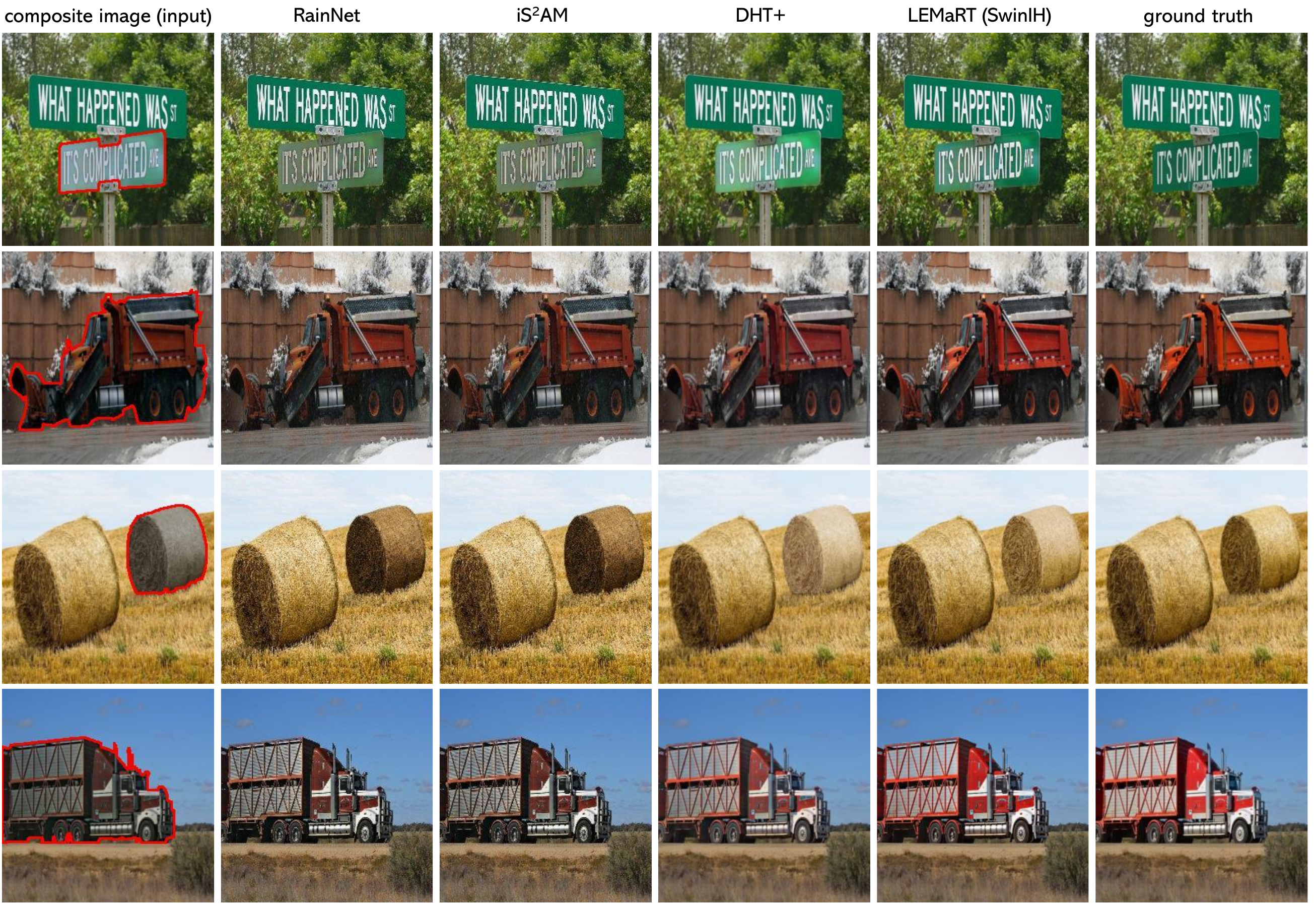}
    \vspace{-1pt}
    \caption{Qualitative comparison between our method, LEMaRT (SwinIH), and three SOTA methods (RainNet\cite{Ling2021RAINNet}, iS$^2$AM\cite{sofiiuk2021foreground}, DHT+\cite{guo2022transformer}) on the iHarmony4 dataset. Compared to other methods, LEMaRT is better at color correction, thanks to the pre-training process during which LEMaRT learns the distribution of photo-realistic images.}
    \vspace{-2pt}
    \label{fig:examples}
\end{figure*}

In Figure \ref{fig:examples}, we compare the harmonized images generated by three SOTA methods, \textit{i.e.}, RainNet, iS$^2$AM, DHT+, and our method, \textit{i.e.}, LEMaRT (SwinIH). We see that LEMaRT can generate photo-realistic images. Compared to other methods, LEMaRT is better at making color corrections, thanks to the pre-training process during which LEMaRT learns the distribution of photo-realistic images.

\vspace{4pt}
\noindent \textbf{b. On RealHM Dataset}
\vspace{1pt}

\noindent In Table \ref{tab:realhm}, we compare the performance of our method, LEMaRT, with multiple SOTA methods on RealHM dataset. We pre-train our model on 120K images from Open Images V6~\cite{kuznetsova2020open} for 22 epochs and then fine-tune our model on iHarmony4 for $1$ epoch with a learning rate of $5.3e^{-3}$. We see that LEMaRT comfortably outperforms DoveNet \cite{Cong2020DoveNet} and S$^{2}$AM \cite{cun2020improving}, and achieves comparable results to SSH \cite{Jiang2021SSH}. A comparison of the harmonized images generated by our LEMaRT method and existing methods can be found in the supplementary materials.

\renewcommand{\arraystretch}{1.0}
\setlength{\tabcolsep}{4pt}
\begin{table}[h]
	\centering
	\begin{tabular}{c||c|c|c|c}
        \specialrule{.1em}{.05em}{.05em} 
		 \rowcolor{mygray} method & DoveNet\cite{Cong2020DoveNet} & S$^{2}$AM\cite{cun2020improving} & SSH\cite{Jiang2021SSH} & LEMaRT \\
        \hline\hline
		MSE $\downarrow$ & 214.1 & 283.3 & 206.9 & \textbf{206.1} \\
		\hline
		PSNR $\uparrow$ & 27.4 & 26.8 & \textbf{27.9} & 27.6 \\
        \specialrule{.1em}{.05em}{.05em} 
	\end{tabular}
    \vspace{-6pt}
	\caption[]{Comparison between our pre-trained model, LEMaRT, and SOTA models on RealHM.}
	\label{tab:realhm}%
	\vspace{-8pt}
\end{table}%

\setlength{\tabcolsep}{6pt}
\begin{table*}[!ht]
	\centering
	\begin{tabular}{c|c||cc|cc|cc|cc|cc}
        \specialrule{.1em}{.05em}{.05em} 
		\rowcolor{mygray} & & \multicolumn{2}{c|}{SwinIH} & \multicolumn{2}{c|}{ViT\cite{dosovitskiy2021an}} & \multicolumn{2}{c|}{ResNet\cite{wang2018pix2pixHD}} & \multicolumn{2}{c|}{HRNet\cite{SunXLW19}} & \multicolumn{2}{c}{HT+\cite{guo2022transformer}} \\
		\rowcolor{mygray} dataset & metric & w/o & w/ & w/o & w/  & w/o & w/ & w/o & w/ & w/o & w/ \\
        \hline \hline
		\multirow{4}[2]{*}{all} & PSNR$\uparrow$ & 37.0 & 39.0 & 35.7 & 38.4 & 34.6 & 36.3 & 33.2 & 35.3 & 37.7 & 38.9 \\
		& MSE$\downarrow$       & 35.5 & 20.9 & 48.2 & 24.1 & 64.4 & 44.2 & 78.5 & 48.4 & 31.4 & 22.3 \\
		& fPSNR$\uparrow$     & 24.5 & 26.6 & 23.1 & 25.9 & 21.9 & 23.4 & 20.6 & 22.7 & 25.1 & 26.3 \\
		& fMSE$\downarrow$      & 386.6 & 250.0 & 499.9 & 282.6 & 645.3 & 459.9 & 811.9 & 510.6 & 342.6 & 266.5 \\
        \specialrule{.1em}{.05em}{.05em} 
	\end{tabular}
    \vspace{-1pt}
	\caption{Effect of pre-training on different image harmonization models, \ie, our SwinIH, {ViT}~\cite{dosovitskiy2021an}, {ResNet}~\cite{wang2018pix2pixHD}, {HRNet}~\cite{SunXLW19} and HT+~\cite{guo2022transformer}. We compare their performance when they are trained on iHarmony4 from scratch (\textit{w/o} columns), and when they are fine-tuned after being pre-trained with LEMaRT using the unlabeled images in MS COCO (\textit{w/} columns).}
	\vspace{-2pt}
	\label{tab:architecture}%
\end{table*}%

\subsection{Ablation Studies}
\noindent We conduct ablation studies to gain insights into various aspects of our method. These aspects include the generalization ability of our LEMaRT method across different network architectures, its efficiency in terms of data and annotation consumption, the design choices of our SwinIH model, and the sensitivity of its performance to the mask generation strategy and the mask size.

\vspace{4pt}
\noindent \textbf{a. Generalization Across Network Architectures}
\vspace{1pt}

\noindent The goal of the first ablation study is to understand the effectiveness of the proposed pre-training method, \textit{i.e.}, LEMaRT, on various network architectures, including vision Transformers and convolutional neural networks (CNNs). Specifically, we adopt five different networks, \textit{i.e.}, SwinIH, {ViT}~\cite{dosovitskiy2021an}, {ResNet}~\cite{wang2018pix2pixHD}, {HRNet}~\cite{SunXLW19} and HT+~\cite{guo2022transformer}. {SwinIH} refers to our model introduced in $\S$~\ref{sec:model}. {ViT} refers to the vision Transformer model that adopts global attention. {ResNet} is a variant of the ResNet generator introduced in pix2pixHD~\cite{wang2018pix2pixHD}. We remove the down sampling operators to make it suitable for image harmonization. We re-implement {HT+}~\cite{guo2022transformer}, a ViT-style Transformer model designed for image harmonization. Our implementation has comparable results ($0.3$ dB higher PSNR, and $0.9$ higher MSE) with those reported in~\cite{guo2022transformer}. We compare the performance of the five networks when they are trained on iHarmony4 from scratch (\textit{w/o} columns in Table \ref{tab:architecture}), and when they are fine-tuned after being pre-trained with LEMaRT (\textit{w/} columns). We train (or fine-tune) SwinIH, ViT, ResNet and HRNet on iHarmony4 for $30$ epochs, and HT+ for $120$ epochs to be consistent with the results reported in~\cite{guo2022transformer}. Other settings are kept the same as the default setting. 

As shown in Table~\ref{tab:architecture}, pre-training on MS COCO with LEMaRT significantly improves performance of the models under study over training from scratch on iHarmony4. Specifically, the performance boost ranges from $1.2$ to $2.7$ dB in terms of PSNR and $1.2$ to $2.8$ dB in terms of fPSNR. In particular, LEMaRT improves the PSNR of our SwinIH model by $2.0$ dB and its MSE by $14.6$. Moreover, LEMaRT is effective not only for models adapted from other vision tasks, but also for those specifically designed for image harmonization, such as HT+~\cite{guo2022transformer}.

\vspace{4pt}
\noindent \textbf{b. Data Efficiency}
\vspace{1pt}
\begin{figure*}
\centering
\includegraphics[width=1.0\linewidth]{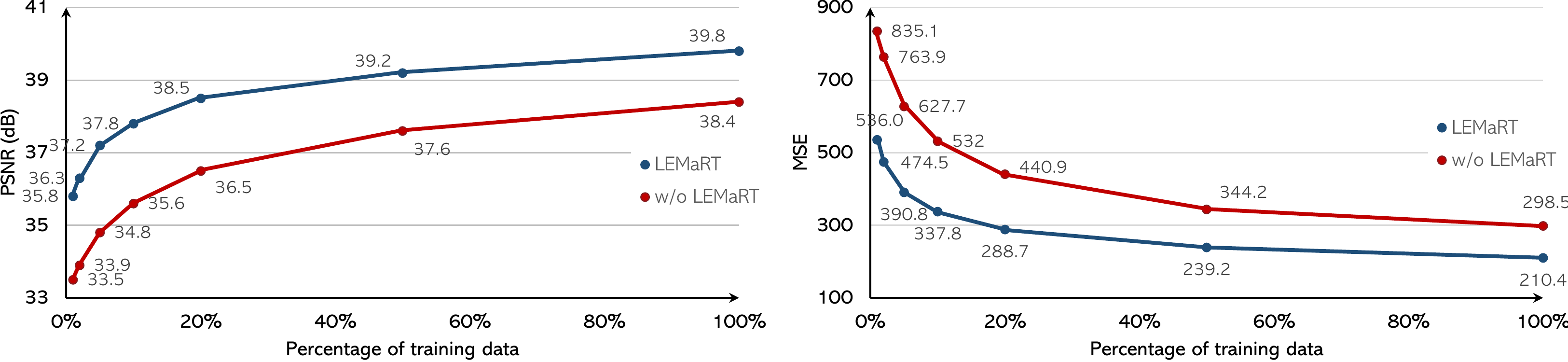}
\caption{Performance of our SwinIH model when it is fine-tuned on a portion ($1$--$50$\%) of the iHarmony4 dataset. One variant is trained from scratch using the all training data from the iHarmony4 dataset (referred to as \textit{w/o LEMaRT}). The other is pre-trained with LEMaRT and then fine-tuned using the all training data from the iHarmony4 dataset (referred to as LEMaRT) .}
\label{fig:percent}
\end{figure*}

\noindent Next, we evaluate the effectiveness of LEMaRT with respect to the amount of fine-tuning data. As before, we pre-train our SwinIH model in two settings: training from scratch on iHarmony4 only and pre-training followed by fine-tuning. For both settings, we vary the amount of fine-tuning data by uniformly sampling between $1$\% and $100$\% of the iHarmony4 training set. 

The results in Figure~\ref{fig:percent} are consistent with the previous section, in the sense that pre-training improves image harmonization accuracy by a large margin (up to $2.4$ dB in terms of PSNR and $299.1$ in terms of MSE) regardless of the amount of fine-tuning data. More importantly, the LEMaRT pre-training scheme is more beneficial to the low data regime than the high data regime. For example, when using no more than $10$\% of the fine-tuning data, the performance boost attributed to pre-training ranges between $2.3$ and $2.4$ dB, whereas the improvement at $100$\% of fine-tuning data declines to $1.4$ dB. We observe a similar trend in the MSE measure, where the MSE improvement drops from around $300.0$ at $1$\% of iHarmony4 training data to less than $90.0$ when using the full training set.

\begin{figure}
    \centering
    \includegraphics[width=0.98\linewidth]{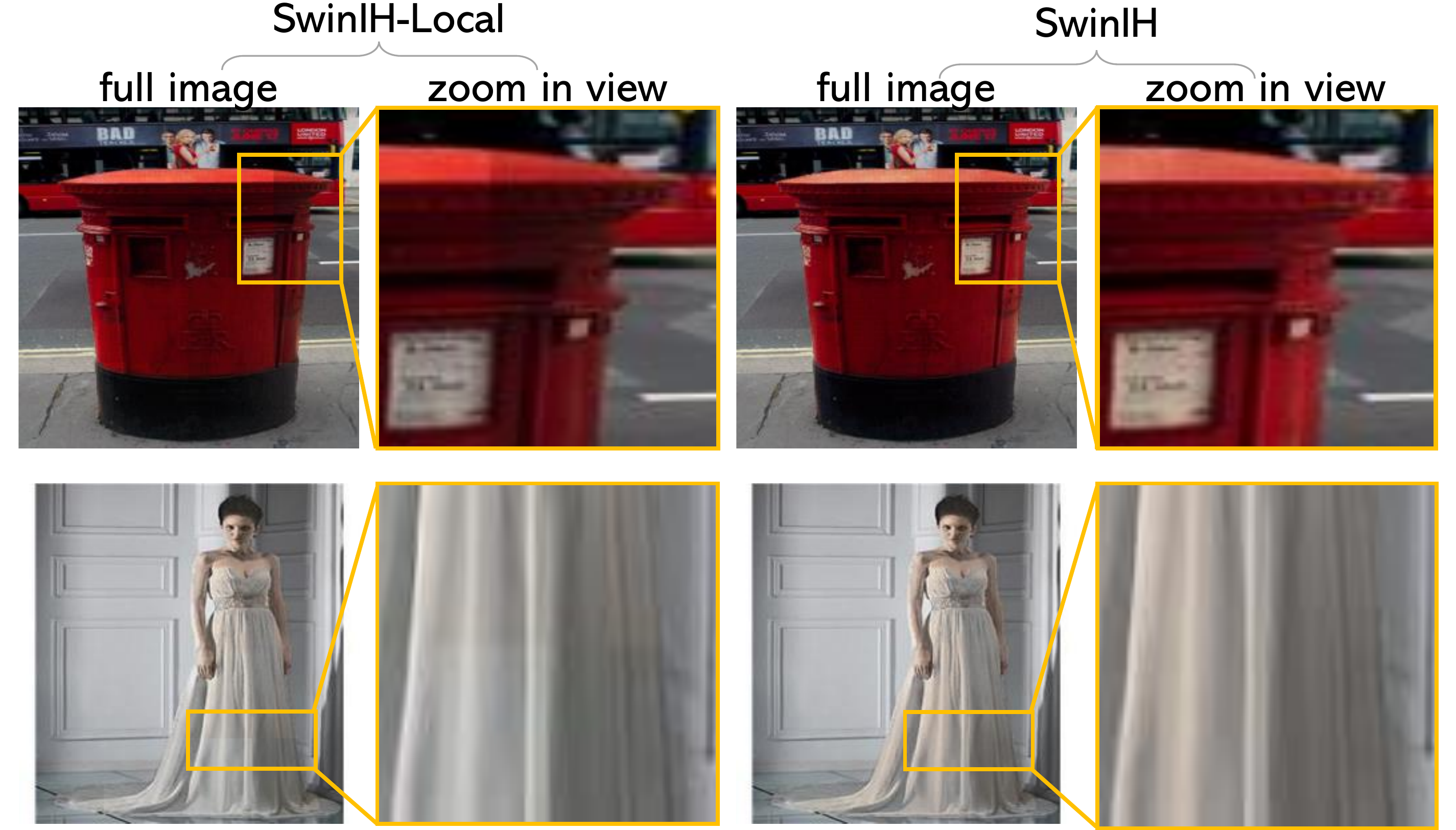}
    \vspace{-2pt}
    \caption{Qualitative comparison between the results of SwinIH-Local and SwinIH. SwinIH-Local occasionally generates images with block-shaped visual artifacts, while SwinIH does not.}
    \vspace{-4pt}
    \label{fig:block}
\end{figure}

\vspace{4pt}
\noindent \textbf{c. Model Design Choices}
\vspace{1pt}
\label{sec:exp-network}

\noindent In this experiment, we pay attention to two design choices of our SwinIH model. The first choice is to maintain the input resolution across all the transformer blocks or to adopt a bottleneck layer similar to encoder-decoder models. The second choice is whether to use efficient local attention, \ie, Swin attention, across all the blocks or to use global attention as well. This choice stems from a visual observation that the Swin attention \textit{occasionally} induces block-shaped visual artifacts in harmonized images, as shown in Figure~\ref{fig:block}. Therefore, it prompts the necessity to modify model architecture to maintain a balance between efficiency and visual quality. 

To gain insights into these aspects, we compare SwinIH, its two variants, \ie, SwinIH-MS, SwinIH-Local, and Swin Transformer (Swin-T)~\cite{Liu2021Swin}. SwinIH is introduced in $\S$~\ref{sec:model}. SwinIH-MS differs from SwinIH in that it first reduces and then enlarges the resolution of feature maps at deeper layers. SwinIH-Local replaces the global attention layer of SwinIH with a Swin attention block. As discussed in $\S$~\ref{sec:model}, Swin-T is composed of Swin attention blocks and uses a Patch Merging layer to reduce the spatial size of feature maps. We add a Patch Splitting layer (does the opposite of a Patch Merging layer) to enlarge the size of feature maps to make it suitable for image harmonization. 

\begin{table}[h]
	\centering
	\begin{tabular}{c|c||cccc}
        \specialrule{.1em}{.05em}{.05em} 
        \rowcolor{mygray} &  & \multicolumn{4}{c}{model} \\
		\rowcolor{mygray} dataset & metric & {SwinIH} & {MS} & {Local} & {Swin-T}\\
        \hline\hline
		\multirow{4}[2]{*}{all} & PSNR$\uparrow$ & 37.0 & {36.3} & {37.0} & 36.1 \\
		& MSE$\downarrow$         & 35.5 & {41.3} & {35.1} & 47.0 \\
		& fPSNR$\uparrow$       & 24.5 & 23.7 & 24.5 & 23.2 \\
		& fMSE$\downarrow$        & 386.6 & {454.5} & {385.2} & 499.3 \\
        \specialrule{.1em}{.05em}{.05em} 
	\end{tabular}
    \vspace{-2pt}
	\caption{Performance comparison of SwinIH, its two variants, \ie, SwinIH-Local (denoted as Local), SwinIH-MS (denoted as MS), and Swin Transformer (denoted as Swin-T) \cite{Liu2021Swin} on iHarmony4.}
	\label{tab:transformer}%
	\vspace{-6pt}
\end{table}%
\setlength{\tabcolsep}{6pt}

As shown in Table~\ref{tab:transformer}, SwinIH significantly outperforms SwinIH-MS and Swin-T across all four metrics, \eg, by $0.7$ dB and $0.9$ dB in terms of PSNR, and by $5.8$ and $11.5$ in terms of MSE, respectively. We hypothesize that the performance drop is caused by the information loss when the resolution of a feature map is reduced. The performance of SwinIH-Local and that of SwinIH are comparable in terms of PSNR and MSE. However, as shown in Figure~\ref{fig:block}, SwinIH produces results that are of higher visual quality. As shown in Figure \ref{fig:block}, SwinIH-Local produces visible block boundaries. This is caused by the shifted window (Swin) attention, which prevents visual tokens at the border of each window to attend to its neighboring visual tokens in adjacent windows. SwinIH is able to remove these block-shaped artifacts. This demonstrates the benefit of using a combination of global and local attention. To maintain high computational and memory efficiency, we only employ it in the last layer of our model.

\vspace{4pt}
\noindent \textbf{d. Mask Generation Strategy}
\vspace{1pt}

\begin{table}[h]
	\centering
	\begin{tabular}{c|c||ccc}
        \specialrule{.1em}{.05em}{.05em} 
        \rowcolor{mygray} & & \multicolumn{3}{c}{mask generation strategy} \\
		\rowcolor{mygray} dataset & metric & random & grid & block \\
        \hline\hline
		\multirow{4}[2]{*}{all} & PSNR$\uparrow$ & 39.0 & 37.1 & 38.6 \\
		& MSE$\downarrow$ & 20.9 & 33.5 & 23.2 \\
		& fPSNR$\uparrow$ & 26.6 & 24.5 & 26.1 \\
		& fMSE$\downarrow$ & 250.0 & 380.3 & 273.8 \\
        \specialrule{.1em}{.05em}{.05em} 
	\end{tabular}
    \vspace{-2pt}
	\caption{Comparison of three mask generation strategies introduced in $\S$~\ref{sec:mask}: \textit{random}, \textit{grid}, \textit{block}, on iHarmony4.
	}
	\vspace{-8pt}
	\label{tab:mask-strategy}%
\end{table}%

\noindent Here we study the sensitivity of harmonization performance with respect to the 
mask generation strategy. To this end, we compare the three strategies  discussed in $\S$~\ref{sec:mask}, \ie, \textit{random}, \textit{grid} and \textit{block}. We measure the performance of our model after being pre-trained on MS COCO and fine-tuned on iHarmony4 for $30$ epochs.

As seen in Table~\ref{tab:mask-strategy}, the performance of \textit{grid} strategy is worse that of the other two strategies. This result is expected, as the \textit{grid} strategy can only transform image patches at specific locations. Therefore, it is not flexible for cases where there are multiple foreground regions or they cover an area larger than a grid cell. To our surprise, the \textit{random} strategy achieves comparable performance to the \textit{block} strategy, which is designed to mimic test cases. This result confirms that there is no need for a special mask generation algorithm that is tuned for the LEMaRT pre-training scheme. In other words, this simplifies the design and broadens the applicability of LEMaRT to new datasets.

\vspace{4pt}
\noindent \textbf{e. Foreground Mask Size}
\vspace{1pt}
\begin{table}[h]
	\centering
	\begin{tabular}{c|c||ccc}
        \specialrule{.1em}{.05em}{.05em} 
        \rowcolor{mygray} & & \multicolumn{3}{c}{mask ratio} \\
		\rowcolor{mygray} dataset & metric & 30\% & 50\% & 70\% \\
        \hline\hline
		\multirow{4}[2]{*}{all} & PSNR$\uparrow$ & 38.8 & 39.0 & 39.0 \\
		& MSE$\downarrow$  & 21.8 & 20.9 & 21.2 \\
		& fPSNR$\uparrow$  & 26.4 & 26.6 & 26.5 \\
		& fMSE$\downarrow$  & 255.0 & 250.0 & 250.5 \\
        \specialrule{.1em}{.05em}{.05em} 
	\end{tabular}
    \vspace{-2pt}
	\caption{Image harmonization metrics corresponding to three different foreground mask ratios on iHarmony4.}
	\vspace{-6pt}
	\label{tab:mask-ratio}%
\end{table}%

\noindent We examine the sensitivity of image harmonization results to the foreground mask size. Here, the foreground size is measured by the ratio of the foreground mask size to the image size. In this experiment, we fine-tune the pre-trained models for $30$ epochs.
In Table \ref{tab:mask-ratio}, we show the quantitative metrics at three different foreground mask ratios, 30\%, 50\% and 70\%. We see that these metrics do not vary significantly between the three ratios. For example, fPSNR varies within a range with $0.2$ dB width and fMSE varies within a range whose width is smaller than $5.0$. This indicates that the size of generated foreground masks does not have significant impact on performance of pre-trained models.

\section{Conclusion}
\noindent In this work, we introduced Label-Efficient Masked Region Transform (LEMaRT), an effective technique of online data generation for self-supervised pre-training of image harmonization models. LEMaRT provides a simple, yet effective way to leverage large-scale unannotated datasets. In addition, we derived a Swin Transformer-based model that is more efficient than ViT-style Transformer networks for image harmonization. Extensive experiments on the iHarmony4 dataset validate the effectiveness of both our  pre-training method and our model. We set a new state of the art for image harmonization, while showing that our pre-training method is much more label-efficient than the existing methods and is consistently applicable to a wide range of network architectures for image harmonization.

{\small
\bibliographystyle{ieee_fullname}
\bibliography{references}

\begin{thebibliography}{10}\itemsep=-1pt

\bibitem{alayrac2022flamingo}
Jean-Baptiste Alayrac, Jeff Donahue, Pauline Luc, Antoine Miech, Iain Barr,
  Yana Hasson, Karel Lenc, Arthur Mensch, Katherine Millican, Malcolm Reynolds,
  et~al.
\newblock Flamingo: a visual language model for few-shot learning.
\newblock {\em Advances in Neural Information Processing Systems},
  35:23716--23736, 2022.

\bibitem{Bao2021BEiT}
Hangbo Bao, Li Dong, Songhao Piao, and Furu Wei.
\newblock {BE}it: {BERT} pre-training of image transformers.
\newblock In {\em International Conference on Learning Representations}, 2022.

\bibitem{Cao2022DIH}
Junyan Cao, Wenyan Cong, Li Niu, Jianfu Zhang, and Liqing Zhang.
\newblock Deep image harmonization by bridging the reality gap.
\newblock In {\em BMVC}, June 2022.

\bibitem{Chen2019Toward}
Bor-Chun Chen and Andrew Kae.
\newblock Toward realistic image compositing with adversarial learning.
\newblock In {\em 2019 IEEE/CVF Conference on Computer Vision and Pattern
  Recognition (CVPR)}, pages 8407--8416, 2019.

\bibitem{chen2020simple}
Ting Chen, Simon Kornblith, Mohammad Norouzi, and Geoffrey Hinton.
\newblock A simple framework for contrastive learning of visual
  representations.
\newblock In {\em ICML}, pages 1597--1607, 2020.

\bibitem{chen2021empirical}
Xinlei Chen, Saining Xie, and Kaiming He.
\newblock An empirical study of training self-supervised vision transformers.
\newblock In {\em Proceedings of the IEEE/CVF International Conference on
  Computer Vision}, pages 9640--9649, 2021.

\bibitem{cong2021bargainnet}
Wenyan Cong, Li Niu, Jianfu Zhang, Jing Liang, and Liqing Zhang.
\newblock Bargainnet: Background-guided domain translation for image
  harmonization.
\newblock In {\em ICME}, pages 1--6, 2021.

\bibitem{Cong2020DoveNet}
Wenyan Cong, Jianfu Zhang, Li Niu, Liu Liu, Zhixin Ling, Weiyuan Li, and Liqing
  Zhang.
\newblock Dovenet: Deep image harmonization via domain verification.
\newblock In {\em CVPR}, 2020.

\bibitem{cun2020improving}
Xiaodong Cun and Chi-Man Pun.
\newblock Improving the harmony of the composite image by spatial-separated
  attention module.
\newblock {\em IEEE Transactions on Image Processing}, 29:4759--4771, 2020.

\bibitem{dosovitskiy2021an}
Alexey Dosovitskiy, Lucas Beyer, Alexander Kolesnikov, Dirk Weissenborn,
  Xiaohua Zhai, Thomas Unterthiner, Mostafa Dehghani, Matthias Minderer, Georg
  Heigold, Sylvain Gelly, Jakob Uszkoreit, and Neil Houlsby.
\newblock An image is worth 16x16 words: Transformers for image recognition at
  scale.
\newblock In {\em International Conference on Learning Representations}, 2021.

\bibitem{Dwibedi2017CutPasteLearn}
Debidatta Dwibedi, Ishan Misra, and Martial Hebert.
\newblock Cut, paste and learn: Surprisingly easy synthesis for instance
  detection.
\newblock In {\em The IEEE International Conference on Computer Vision (ICCV)},
  Oct 2017.

\bibitem{Ghiasi2021SimpleCopyPaste}
Golnaz Ghiasi, Yin Cui, Aravind Srinivas, Rui Qian, Tsung-Yi Lin, Ekin~D.
  Cubuk, Quoc~V. Le, and Barret Zoph.
\newblock Simple copy-paste is a strong data augmentation method for instance
  segmentation.
\newblock In {\em Proceedings of the IEEE/CVF Conference on Computer Vision and
  Pattern Recognition (CVPR)}, pages 2918--2928, June 2021.

\bibitem{guo2022transformer}
Zonghui Guo, Zhaorui Gu, Bing Zheng, Junyu Dong, and Haiyong Zheng.
\newblock Transformer for image harmonization and beyond.
\newblock {\em IEEE Transactions on Pattern Analysis and Machine Intelligence},
  2022.

\bibitem{Guo_2021_ICCV}
Zonghui Guo, Dongsheng Guo, Haiyong Zheng, Zhaorui Gu, Bing Zheng, and Junyu
  Dong.
\newblock Image harmonization with transformer.
\newblock In {\em ICCV}, pages 14870--14879, October 2021.

\bibitem{Guo2021Intrinsic}
Zonghui Guo, Haiyong Zheng, Yufeng Jiang, Zhaorui Gu, and Bing Zheng.
\newblock Intrinsic image harmonization.
\newblock In {\em Proceedings of the IEEE/CVF Conference on Computer Vision and
  Pattern Recognition (CVPR)}, pages 16367--16376, June 2021.

\bibitem{Hang2022SCSCo}
Yucheng Hang, Bin Xia, Wenming Yang, and Qingmin Liao.
\newblock Scs-co: Self-consistent style contrastive learning for image
  harmonization.
\newblock In {\em 2022 IEEE/CVF Conference on Computer Vision and Pattern
  Recognition (CVPR)}, pages 19678--19687, 2022.

\bibitem{He2021MAE}
Kaiming He, Xinlei Chen, Saining Xie, Yanghao Li, Piotr Doll{\'{a}}r, and
  Ross~B. Girshick.
\newblock Masked autoencoders are scalable vision learners.
\newblock In {\em {IEEE/CVF} Conference on Computer Vision and Pattern
  Recognition, {CVPR} 2022, New Orleans, LA, USA, June 18-24, 2022}, pages
  15979--15988. {IEEE}, 2022.

\bibitem{he2020momentum}
Kaiming He, Haoqi Fan, Yuxin Wu, Saining Xie, and Ross Girshick.
\newblock Momentum contrast for unsupervised visual representation learning.
\newblock In {\em CVPR}, pages 9729--9738, 2020.

\bibitem{Jia2021ALIGN}
Chao Jia, Yinfei Yang, Ye Xia, Yi-Ting Chen, Zarana Parekh, Hieu Pham, Quoc Le,
  Yun-Hsuan Sung, Zhen Li, and Tom Duerig.
\newblock Scaling up visual and vision-language representation learning with
  noisy text supervision.
\newblock In Marina Meila and Tong Zhang, editors, {\em Proceedings of the 38th
  International Conference on Machine Learning}, volume 139 of {\em Proceedings
  of Machine Learning Research}, pages 4904--4916. PMLR, 18--24 Jul 2021.

\bibitem{Jiang2021SSH}
Yifan Jiang, He Zhang, Jianming Zhang, Yilin Wang, Zhe Lin, Kalyan Sunkavalli,
  Simon Chen, Sohrab Amirghodsi, Sarah Kong, and Zhangyang Wang.
\newblock Ssh: A self-supervised framework for image harmonization.
\newblock In {\em Proceedings of the IEEE/CVF International Conference on
  Computer Vision}, pages 4832--4841, 2021.

\bibitem{kuznetsova2020open}
Alina Kuznetsova, Hassan Rom, Neil Alldrin, Jasper Uijlings, Ivan Krasin, Jordi
  Pont-Tuset, Shahab Kamali, Stefan Popov, Matteo Malloci, Alexander
  Kolesnikov, et~al.
\newblock The open images dataset v4: Unified image classification, object
  detection, and visual relationship detection at scale.
\newblock {\em International Journal of Computer Vision}, 128(7):1956--1981,
  2020.

\bibitem{Lalonde2007ColorCompatibility}
Jean-Francois Lalonde and Alexei~A. Efros.
\newblock Using color compatibility for assessing image realism.
\newblock In {\em 2007 IEEE 11th International Conference on Computer Vision},
  pages 1--8, 2007.

\bibitem{Lee2019VidIntoVid}
Donghoon Lee, Tomas Pfister, and Ming-Hsuan Yang.
\newblock Inserting videos into videos.
\newblock In {\em Proceedings of the IEEE/CVF Conference on Computer Vision and
  Pattern Recognition (CVPR)}, June 2019.

\bibitem{Liang2022RegionCL}
Jingtang Liang and Chi-Man Pun.
\newblock Image harmonization with region-wise contrastive learning, 2022.

\bibitem{Lin2014}
Tsung-Yi Lin, Michael Maire, Serge Belongie, James Hays, Pietro Perona, Deva
  Ramanan, Piotr Dollar, and C~Lawrence Zitnick.
\newblock Deep high-resolution representation learning for human pose
  estimation.
\newblock In {\em ECCV}, 2014.

\bibitem{Ling2021RAINNet}
Jun Ling, Han Xue, Li Song, Rong Xie, and Xiao Gu.
\newblock Region-aware adaptive instance normalization for image harmonization.
\newblock In {\em Proceedings of the IEEE/CVF Conference on Computer Vision and
  Pattern Recognition (CVPR)}, pages 9361--9370, June 2021.

\bibitem{Liu2021Swin}
Ze Liu, Yutong Lin, Yue Cao, Han Hu, Yixuan Wei, Zheng Zhang, Stephen Lin, and
  Baining Guo.
\newblock Swin transformer: Hierarchical vision transformer using shifted
  windows.
\newblock In {\em Proceedings of the IEEE/CVF International Conference on
  Computer Vision (ICCV)}, 2021.

\bibitem{loshchilov2018fixing}
Ilya Loshchilov and Frank Hutter.
\newblock Fixing weight decay regularization in adam, 2018.

\bibitem{Radford2013CLIP}
Alec Radford, Jong~Wook Kim, Chris Hallacy, Aditya Ramesh, Gabriel Goh,
  Sandhini Agarwal, Girish Sastry, Amanda Askell, Pamela Mishkin, Jack Clark,
  Gretchen Krueger, and Ilya Sutskever.
\newblock Learning transferable visual models from natural language
  supervision.
\newblock {\em CoRR}, abs/2103.00020, 2021.

\bibitem{sofiiuk2021foreground}
Konstantin Sofiiuk, Polina Popenova, and Anton Konushin.
\newblock Foreground-aware semantic representations for image harmonization.
\newblock In {\em WACV}, pages 1620--1629, 2021.

\bibitem{Song2020Illumination}
Shuangbing Song, Fan Zhong, Xueying Qin, and Changhe Tu.
\newblock Illumination harmonization with gray mean scale.
\newblock In {\em Advances in Computer Graphics: 37th Computer Graphics
  International Conference, CGI 2020, Geneva, Switzerland, October 20–23,
  2020, Proceedings}, page 193–205, Berlin, Heidelberg, 2020.
  Springer-Verlag.

\bibitem{SunXLW19}
Ke Sun, Bin Xiao, Dong Liu, and Jingdong Wang.
\newblock Deep high-resolution representation learning for human pose
  estimation.
\newblock In {\em CVPR}, 2019.

\bibitem{Sunkavalli2010MultiscaleIH}
Kalyan Sunkavalli, Micah~K. Johnson, Wojciech Matusik, and Hanspeter Pfister.
\newblock Multi-scale image harmonization.
\newblock {\em ACM Trans. Graph.}, 29(4), jul 2010.

\bibitem{tsai2017deep}
Yi-Hsuan Tsai, Xiaohui Shen, Zhe Lin, Kalyan Sunkavalli, Xin Lu, and Ming-Hsuan
  Yang.
\newblock Deep image harmonization.
\newblock In {\em CVPR}, pages 3789--3797, 2017.

\bibitem{Wang2019PSIS}
Hao Wang, Qilong Wang, Fan Yang, Weiqi Zhang, and Wangmeng Zuo.
\newblock Data augmentation for object detection via progressive and selective
  instance-switching.
\newblock {\em CoRR}, abs/1906.00358, 2019.

\bibitem{wang2022omnivl}
Junke Wang, Dongdong Chen, Zuxuan Wu, Chong Luo, Luowei Zhou, Yucheng Zhao,
  Yujia Xie, Ce Liu, Yu-Gang Jiang, and Lu Yuan.
\newblock Omnivl: One foundation model for image-language and video-language
  tasks.
\newblock {\em arXiv preprint arXiv:2209.07526}, 2022.

\bibitem{Wang2018Vid2Vid}
Ting-Chun Wang, Ming-Yu Liu, Jun-Yan Zhu, Guilin Liu, Andrew Tao, Jan Kautz,
  and Bryan Catanzaro.
\newblock Video-to-video synthesis.
\newblock In {\em Advances in Neural Information Processing Systems (NeurIPS)},
  2018.

\bibitem{wang2018pix2pixHD}
Ting-Chun Wang, Ming-Yu Liu, Jun-Yan Zhu, Andrew Tao, Jan Kautz, and Bryan
  Catanzaro.
\newblock High-resolution image synthesis and semantic manipulation with
  conditional gans.
\newblock In {\em Proceedings of the IEEE Conference on Computer Vision and
  Pattern Recognition}, 2018.

\bibitem{Xue2012Understanding}
Su Xue, Aseem Agarwala, Julie Dorsey, and Holly Rushmeier.
\newblock Understanding and improving the realism of image composites.
\newblock {\em ACM Trans. Graph.}, 31(4), July 2012.

\bibitem{yuan2021florence}
Lu Yuan, Dongdong Chen, Yi-Ling Chen, Noel Codella, Xiyang Dai, Jianfeng Gao,
  Houdong Hu, Xuedong Huang, Boxin Li, Chunyuan Li, et~al.
\newblock Florence: A new foundation model for computer vision.
\newblock {\em arXiv preprint arXiv:2111.11432}, 2021.

\bibitem{zhong2022regionclip}
Yiwu Zhong, Jianwei Yang, Pengchuan Zhang, Chunyuan Li, Noel Codella,
  Liunian~Harold Li, Luowei Zhou, Xiyang Dai, Lu Yuan, Yin Li, et~al.
\newblock Regionclip: Region-based language-image pretraining.
\newblock In {\em Proceedings of the IEEE/CVF Conference on Computer Vision and
  Pattern Recognition}, pages 16793--16803, 2022.

\bibitem{Zhou2019DPR}
Hao Zhou, Sunil Hadap, Kalyan Sunkavalli, and David~W. Jacobs.
\newblock Deep single portrait image relighting.
\newblock In {\em International Conference on Computer Vision (ICCV)}, 2019.

\bibitem{zhou2021uc2}
Mingyang Zhou, Luowei Zhou, Shuohang Wang, Yu Cheng, Linjie Li, Zhou Yu, and
  Jingjing Liu.
\newblock Uc2: Universal cross-lingual cross-modal vision-and-language
  pre-training.
\newblock In {\em Proceedings of the IEEE/CVF Conference on Computer Vision and
  Pattern Recognition}, pages 4155--4165, 2021.

\bibitem{Zhu2020Indomain}
Jiapeng Zhu, Yujun Shen, Deli Zhao, and Bolei Zhou.
\newblock In-domain gan inversion for real image editing.
\newblock In {\em Proceedings of European Conference on Computer Vision
  (ECCV)}, 2020.

\bibitem{Zhu2015Learning}
Jun-Yan Zhu, Philipp Kr{\"a}henb{\"u}hl, Eli Shechtman, and Alexei~A. Efros.
\newblock Learning a discriminative model for the perception of realism in
  composite images.
\newblock In {\em IEEE International Conference on Computer Vision (ICCV)},
  2015.

\end{thebibliography}


\begin{thebibliography}{10}\itemsep=-1pt

\bibitem{cong2021bargainnet}
Wenyan Cong, Li Niu, Jianfu Zhang, Jing Liang, and Liqing Zhang.
\newblock Bargainnet: Background-guided domain translation for image
  harmonization.
\newblock In {\em ICME}, pages 1--6, 2021.

\bibitem{Cong2022HighResolutionIH}
Wenyan Cong, Xinhao Tao, Li Niu, Jing Liang, Xuesong Gao, Qihao Sun, and Liqing
  Zhang.
\newblock High-resolution image harmonization via collaborative dual
  transformations.
\newblock In {\em 2022 IEEE/CVF Conference on Computer Vision and Pattern
  Recognition (CVPR)}, pages 18449--18458, 2022.

\bibitem{Cong2020DoveNet}
Wenyan Cong, Jianfu Zhang, Li Niu, Liu Liu, Zhixin Ling, Weiyuan Li, and Liqing
  Zhang.
\newblock Dovenet: Deep image harmonization via domain verification.
\newblock In {\em CVPR}, 2020.

\bibitem{cun2020improving}
Xiaodong Cun and Chi-Man Pun.
\newblock Improving the harmony of the composite image by spatial-separated
  attention module.
\newblock {\em IEEE Transactions on Image Processing}, 29:4759--4771, 2020.

\bibitem{guo2022transformer}
Zonghui Guo, Zhaorui Gu, Bing Zheng, Junyu Dong, and Haiyong Zheng.
\newblock Transformer for image harmonization and beyond.
\newblock {\em IEEE Transactions on Pattern Analysis and Machine Intelligence},
  2022.

\bibitem{Guo2021Intrinsic}
Zonghui Guo, Haiyong Zheng, Yufeng Jiang, Zhaorui Gu, and Bing Zheng.
\newblock Intrinsic image harmonization.
\newblock In {\em Proceedings of the IEEE/CVF Conference on Computer Vision and
  Pattern Recognition (CVPR)}, pages 16367--16376, June 2021.

\bibitem{Hang2022SCSCo}
Yucheng Hang, Bin Xia, Wenming Yang, and Qingmin Liao.
\newblock Scs-co: Self-consistent style contrastive learning for image
  harmonization.
\newblock In {\em 2022 IEEE/CVF Conference on Computer Vision and Pattern
  Recognition (CVPR)}, pages 19678--19687, 2022.

\bibitem{Ke2022Harmonizer}
Zhanghan Ke, Chunyi Sun, Lei Zhu, Ke Xu, and Rynson W.~H. Lau.
\newblock Harmonizer: Learning to perform white-box image and video
  harmonization.
\newblock In {\em ECCV}, 2022.

\bibitem{Liang2022SpatialSeparated}
Jingtang Liang, Xiaodong Cun, Chi-Man Pun, and Jue Wang.
\newblock Spatial-separated curve rendering network for efficient and
  high-resolution image harmonization.
\newblock In {\em ECCV}, 2022.

\bibitem{Ling2021RAINNet}
Jun Ling, Han Xue, Li Song, Rong Xie, and Xiao Gu.
\newblock Region-aware adaptive instance normalization for image harmonization.
\newblock In {\em Proceedings of the IEEE/CVF Conference on Computer Vision and
  Pattern Recognition (CVPR)}, pages 9361--9370, June 2021.

\bibitem{peng2022frih}
Jinlong Peng, Zekun Luo, Liang Liu, Boshen Zhang, Tao Wang, Yabiao Wang, Ying
  Tai, Chengjie Wang, and Weiyao Lin.
\newblock Frih: Fine-grained region-aware image harmonization.
\newblock {\em arXiv preprint arXiv:2205.06448}, 2022.

\bibitem{sofiiuk2021foreground}
Konstantin Sofiiuk, Polina Popenova, and Anton Konushin.
\newblock Foreground-aware semantic representations for image harmonization.
\newblock In {\em WACV}, pages 1620--1629, 2021.

\bibitem{tsai2017deep}
Yi-Hsuan Tsai, Xiaohui Shen, Zhe Lin, Kalyan Sunkavalli, Xin Lu, and Ming-Hsuan
  Yang.
\newblock Deep image harmonization.
\newblock In {\em CVPR}, pages 3789--3797, 2017.

\end{thebibliography}
}
\end{document}


\title{Supplementary Material \\ LEMaRT: Label-Efficient Masked Region Transform for Image Harmonization}

\author{Sheng Liu \quad Cong Phuoc Huynh \quad Cong Chen \quad Maxim Arap \quad Raffay Hamid\\
Amazon Prime Video\\
{\tt\small \{shenlu, conghuyn, checongt, maxarap, raffay\}@amazon.com}
}
\maketitle


\section{Qualitative Examples}
\label{sec:qualitative}

\noindent We present additional qualitative examples of the output of our method (LEMaRT [SwinIH]) and three SOTA methods (RainNet\cite{Ling2021RAINNet}, iS$^2$AM\cite{sofiiuk2021foreground}, DHT+\cite{guo2022transformer}) on the iHarmony4 dataset in Figure \ref{fig:examples1} and Figure \ref{fig:examples2}. Similar to what we observe in Figure 5 of the main paper, our method is better at color correction. The images generated by our method have more natural colors and are closer to the ground truth images. We also provide qualitative examples of the output of our method (LEMaRT [SwinIH]), iS$^2$AM\cite{sofiiuk2021foreground} and DHT+\cite{guo2022transformer} on the RealHM dataset in Figure \ref{fig:examples3}. We see from the first five examples that our method better harmonizes composite images. We show a controversial example in the last row. Different people may have different opinions regarding which harmonized image looks more natural.

\section{Details of Data Generation Pipeline}
\label{sec:transformation}

\noindent The data generation pipeline of LEMaRT uses ten different transformations to generate the data for pre-training. The transformations include adjustments to brightness, contrast, hue, saturation and sharpness as well as blurring, deblurring, auto contrast, equalization and posterization. The first five transformations adjust the brightness, contrast, hue, saturation and sharpness of an image by a factor of $c$, respectively, $c \in [0.2, 1.8]$ for brightness adjustment; $c \in [0.3, 1.7]$ for contrast adjustment; $c \in [0.7, 1.3]$ for hue adjustment; $c \in [0.5, 1.5]$ for saturation adjustment; $c \in [0.0, 2.0]$ for sharpness adjustment. We sample $c$ uniformly. For blurring, Gaussian blur with kernel size $(k_1, k_2)$ ($k_1 \in [3, 9], k_2 \in [5, 11]$) is applied. For deblurring, we apply Gaussian blur to an image. The \textit{blurred image} is treated as the original image and the unblurred image is treated as the transformed image. Auto contrast maximizes the contrast of an image by remapping its pixel values so that the lowest value becomes $0$ and the highest value becomes $255$. Equalization adjusts the histogram of an image so that the histogram of the output image has a uniform grayscale distribution. Posterization reduces the number of bits for each color channel of an image to $n$ bits. $n$ is uniformly sampled from $\{1, 2, 3, 4, 5, 6\}$.

\begin{table}[h]
	\centering
	\begin{tabular}{c|c||cc}
        \specialrule{.1em}{.05em}{.05em} 
        \rowcolor{mygray} & & \multicolumn{2}{c}{ transformation diversity} \\
		\rowcolor{mygray} dataset & metric & standard & less \\
        \hline\hline
		\multirow{4}[2]{*}{all} & PSNR$\uparrow$ & 39.0 & 38.2 \\
		& MSE$\downarrow$ & 20.9 & 26.0 \\
		& fPSNR$\uparrow$ & 26.6 & 25.7 \\
		& fMSE$\downarrow$ & 250.0 & 301.2 \\
        \specialrule{.1em}{.05em}{.05em} 
	\end{tabular}
    \vspace{-2pt}
	\caption{Comparison of the performance of our method pre-trained using different transformations.}
	\vspace{-8pt}
	\label{tab:transformation}%
\end{table}%

In Figure \ref{fig:pretrain-examples}, we present additional examples that show data generated for pre-training, \ie, the transformed images, the masks and the composite images, and the output of a pre-trained LEMaRT model given the generated data (please refer to Figure 2 of the main paper for an illustration of the data generation pipeline of LEMaRT). The seven different transformations used to perturb the original images are attached to the transformed images. We can see that, after being pre-trained, our LEMaRT model can harmonize the composite images that are generated using a variety of transformations, \eg, brightness adjustment, posterization, blurring. The results show that the our LEMaRT model learns to handle different factors that cause appearance mismatch between the foreground and the surrounding background. 

To understand the impact of the diversity of transformations to the performance of our model (LEMaRT [SwinIH]), in Table \ref{tab:transformation}, we compare the performance of our model pre-trained with the set of transformations introduced above (denoted as standard) and a set of transformations with less diversity (denoted as less). Specifically, we halve the range of the factor $c$ that controls the diversity of the five transformations which adjust the brightness, contrast, hue, saturation and sharpness of an image, \ie, $c \in [0.6, 1.4]$ for brightness adjustment; $c \in [0.65, 1.35]$ for contrast adjustment; $c \in [0.85, 1.15]$ for hue adjustment; $c \in [0.75, 1.25]$ for saturation adjustment; $c \in [0.5, 1.0]$ for sharpness adjustment. We sample $c$ uniformly. For blurring, Gaussian blur with kernel size $(k_1, k_2)$ ($k_1 \in [2, 5], k_2 \in [2, 5]$) is applied. We do not use equalization. We see from Table \ref{tab:transformation} that pre-training our model using transformations with less diversity results in $0.8$ dB drop in PSNR and $5.1$ increase in MSE. This indicates that the diversity of the transformations has direct influence on the performance of our model. However, empirically, we find that increasing the diversity of the transformations further does not lead to better performance. A possible reason is that samples created by those transformations are so unnatural that they seldom appear in real world.

\section{Comparison with SOTA Methods}

\noindent We compare our method, LEMaRT [SwinIH], with SOTA methods on iHarmony4. The results are shown Table \ref{tab:sota-ihd}. Table \ref{tab:sota-ihd} differs from the Table 1 of the main paper only in that it shows all four metrics, \ie, PSNR, MSE, fPSNR, fMSE (due to space constraints, the Table 1 of the main paper does not show fPSNR and fMSE). Similar to what we observe from the Table 1 of the main paper, our method consistently outperforms other methods across the two additional metrics (fPSNR and fMSE) on all subsets of iHarmony4. Our method achieves a fPSNR of $27.2$ dB, which is $1.3$ dB higher than the previous best method. The fMSE of our method is $213.3$, which is $35.6$ lower ($14.3$\% relative improvement) than the previous best method~\cite{Hang2022SCSCo}.

In Table \ref{tab:sota-eccv}, we further compare our method with four additional image harmonization methods \cite{peng2022frih,Cong2022HighResolutionIH,Liang2022SpatialSeparated,Ke2022Harmonizer} for completeness. \cite{Liang2022SpatialSeparated,Ke2022Harmonizer} are published in ECCV'22. \cite{peng2022frih} is an arXiv paper and \cite{Cong2022HighResolutionIH} is a CVPR'22 paper. These four methods \textit{underperform} SCS-Co\cite{Hang2022SCSCo} with which we compare our method in Table 1 of the main paper. We only present PSNR and MSE as all four methods do not report fPSNR and three of them, \ie, \cite{peng2022frih,Cong2022HighResolutionIH,Liang2022SpatialSeparated}, do not report fMSE. We see that our method outperforms all four methods. Our method achieves a PSNR of $39.8$ dB which is $1.6$ dB higher than \cite{peng2022frih}, \ie, the best of the four methods. The MSE of our method is $7.2$ lower ($30.0$\% relative improvement) than \cite{peng2022frih}. Harmonizer also adopts a perturbation-reconstruction strategy for training data generation. While Harmonizer \cite{Ke2022Harmonizer} applies transformations to perturb the manually labeled foreground, LEMaRT perturbs regions specified by automatically generated masks. LEMaRT generates training data automatically using the plentiful supply of unlabeled data. We see that the MSE of our method is $8.8$ higher than that of Harmonizer. It is likely that this is caused by a few images on which our method performs poorly (largest MSE over $1200.0$, more than 27 times of the average). As PSNR is in log space, the images on which our method performs poorly has less influence to PSNR than MSE.

\section{Cause of Block-shaped Artifacts}

\begin{figure}[h]
    \centering
    \includegraphics[width=0.60\linewidth]{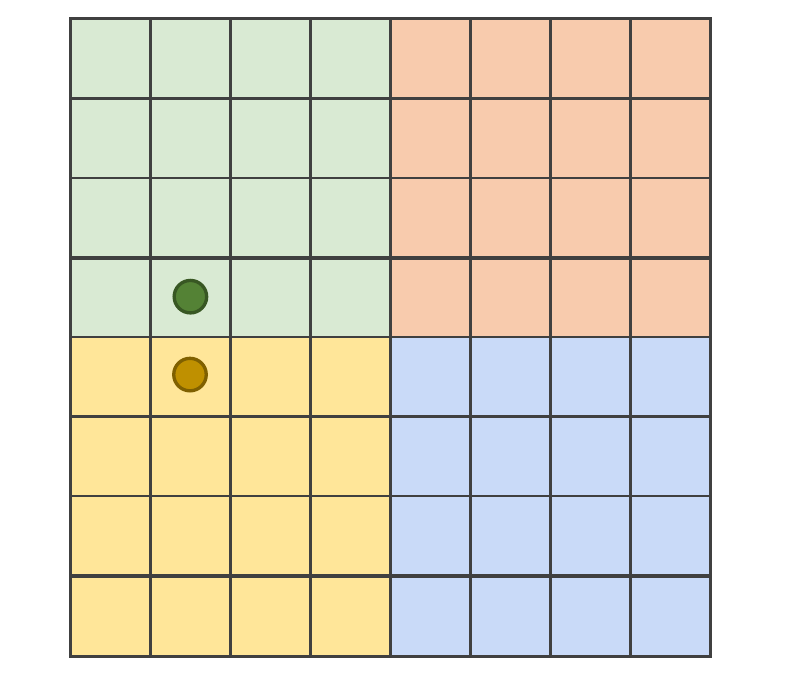}
    \vspace{-1pt}
    \caption{Illustration of the shifted window (Swin) attention which performs self-attention within local windows (window size is $4 \times 4$). Visual tokens within the same window (shown in the same color) can attend to each other, but cannot attend to their neighboring tokens in other windows. For example, the two tokens that contain circles cannot attend to each other even though they are next to each other. Hence, only using Swin attention \textit{may} cause block-shaped artifacts (shown in Figure 7 of the main paper).}
    \vspace{-2pt}
    \label{fig:block}
\end{figure}

\begin{figure*}
    \centering
    \includegraphics[width=0.80\linewidth]{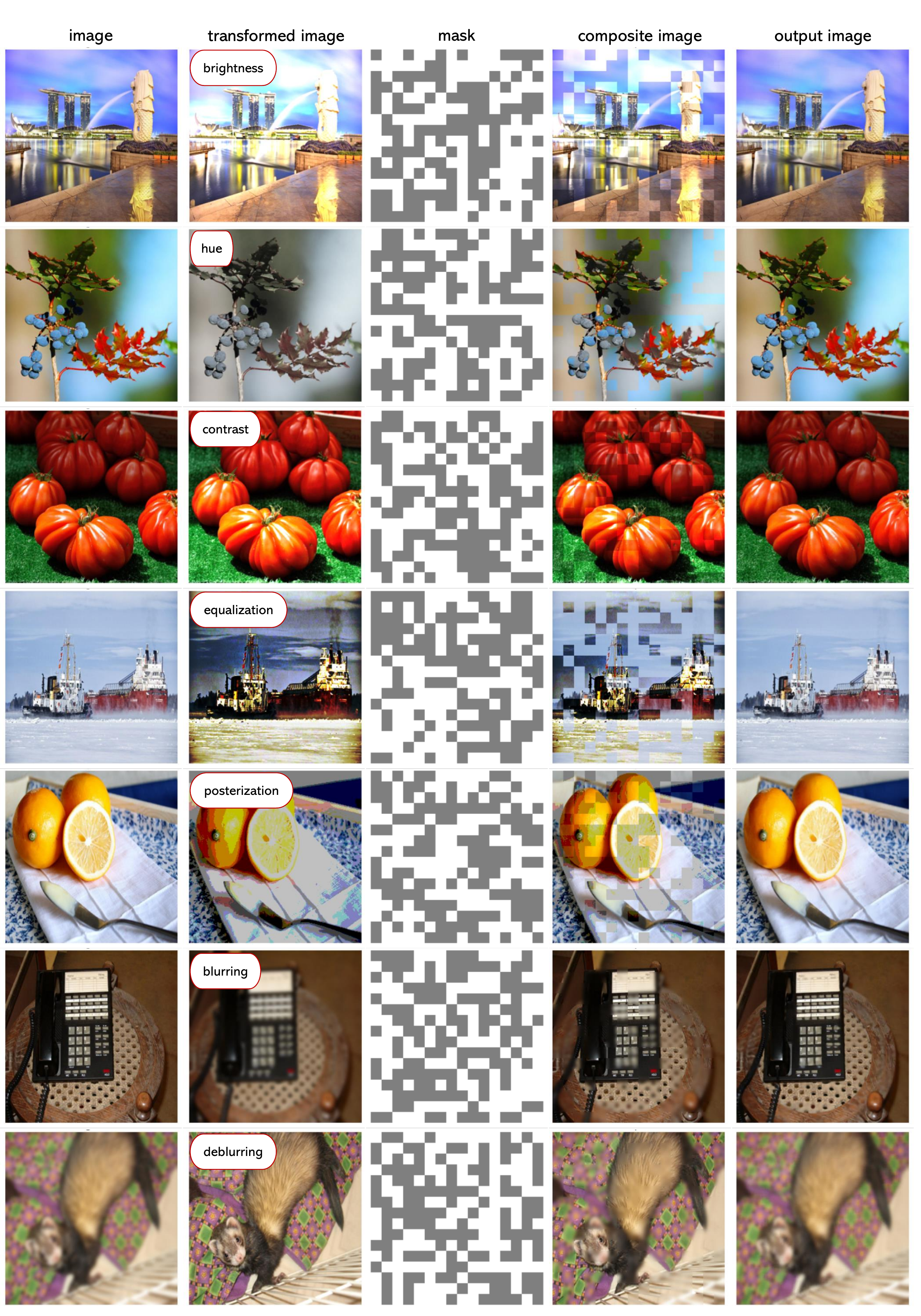}
    \vspace{-1pt}
    \caption{Qualitative examples that show data generated for pre-training and the output of a pre-trained LEMaRT model given the generated data. We apply brightness adjustment (row 1), hue adjustment (row 2), contrast adjustment (row 3), equalization (row 4), posterization (row 5), blurring (row 6), deblurring (row 7) to generate the transformed images shown in the second column.}
    \vspace{-2pt}
    \label{fig:pretrain-examples}
\end{figure*}

\begin{figure*}
    \centering
    \includegraphics[width=0.985\linewidth]{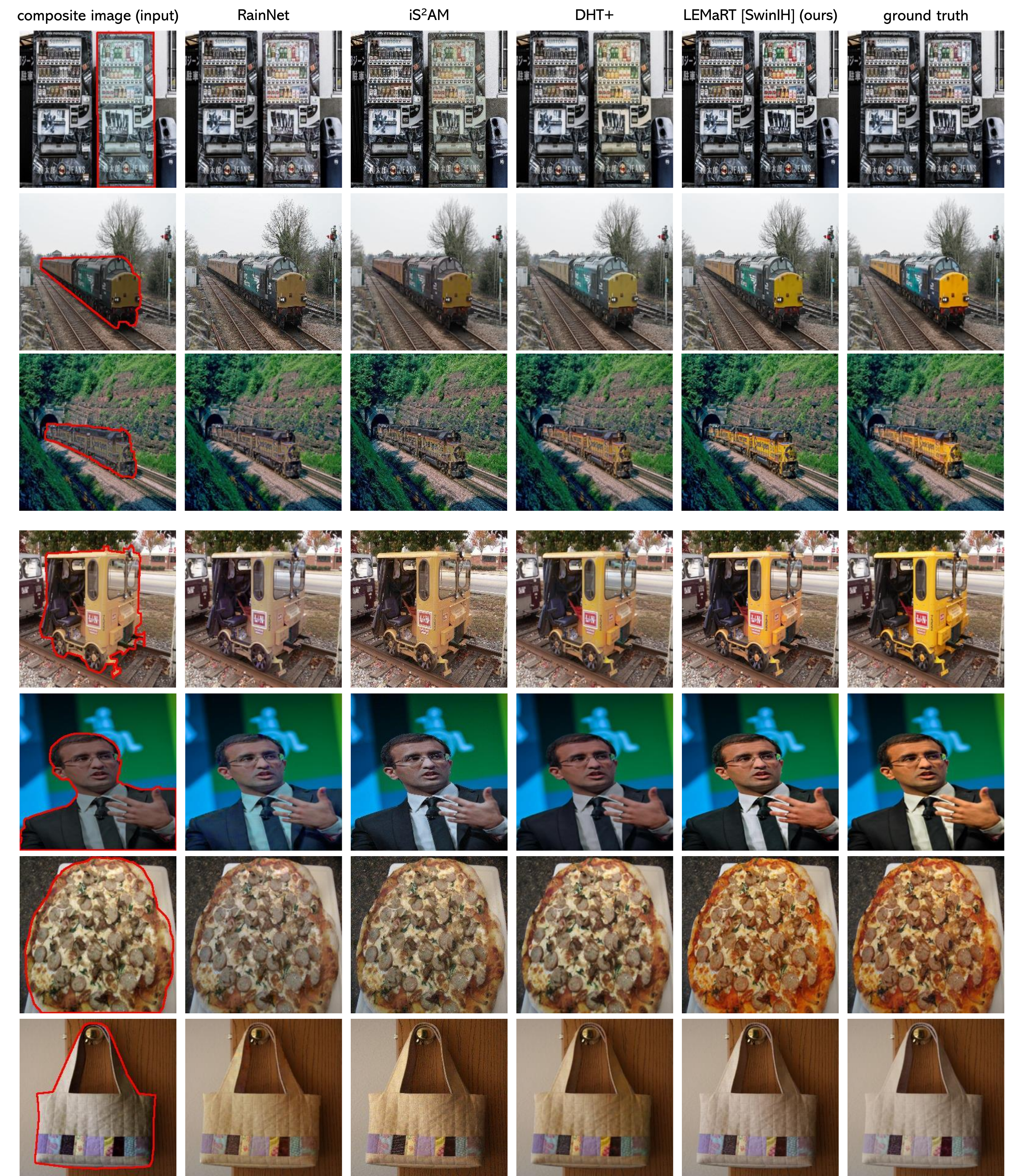}
    \vspace{-1pt}
    \caption{Qualitative comparison between our method (LEMaRT [SwinIH]) and three SOTA methods (RainNet\cite{Ling2021RAINNet}, iS$^2$AM\cite{sofiiuk2021foreground}, DHT+\cite{guo2022transformer}) on iHarmony4. Compared to other methods, LEMaRT is better at color correction, thanks to the pre-training process during which LEMaRT learns the distribution of photo-realistic images.}
    \vspace{-2pt}
    \label{fig:examples1}
\end{figure*}

\begin{figure*}
    \centering
    \includegraphics[width=0.95\linewidth]{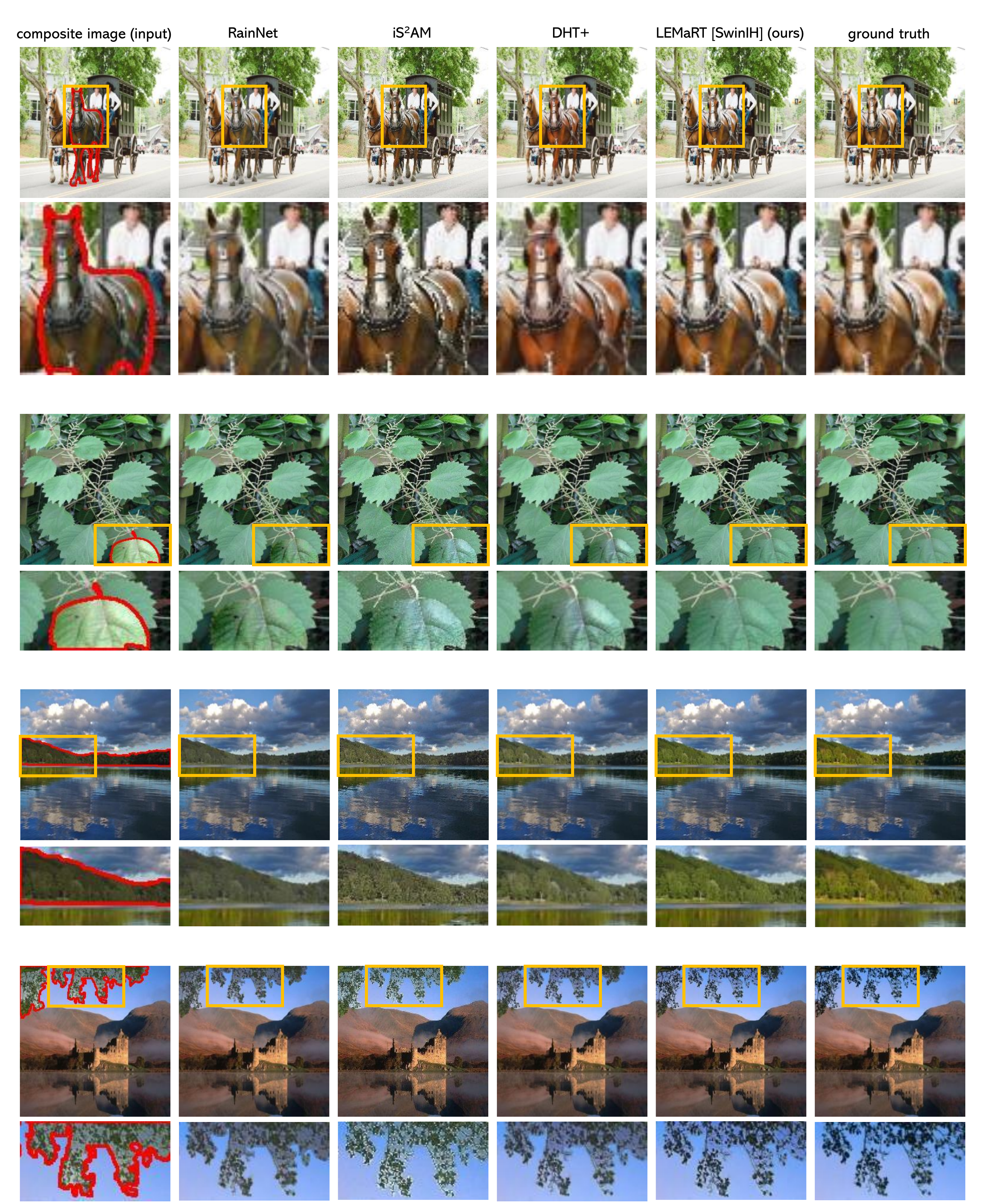}
    \vspace{-1pt}
    \caption{Qualitative comparison between our method, LEMaRT [SwinIH], and three SOTA methods (RainNet\cite{Ling2021RAINNet}, iS$^2$AM\cite{sofiiuk2021foreground}, DHT+\cite{guo2022transformer}) on  iHarmony4. We provide zoom-in views of regions in yellow rectangles. In the first example, the color of the horse in our image is more natural and closer to that of the horse in the ground truth image than other images. We see from the second example that the texture and the color of the leaf in our image are in harmony with those of the background. In the third example, the color of the mountain and its reflection are better aligned in our image than other images. We see from the forth example that our method can better harmonize subtle structures, \ie, small leaves and thin branches, than other methods.}
    \vspace{-2pt}
    \label{fig:examples2}
\end{figure*}

\begin{figure*}
    \centering
    \includegraphics[width=0.95\linewidth]{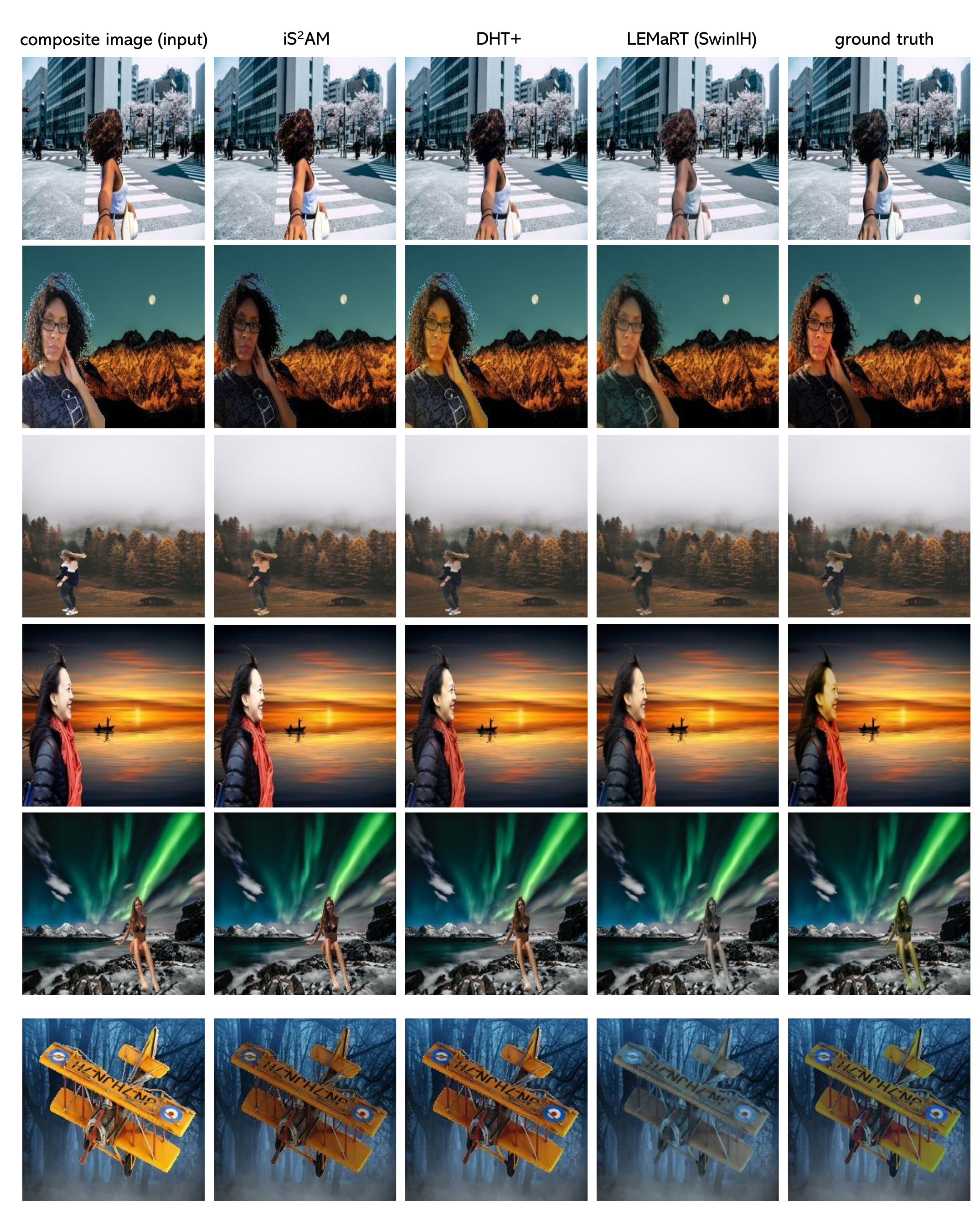}
    \vspace{-1pt}
    \caption{Qualitative comparison between our method, LEMaRT [SwinIH], and two SOTA methods (iS$^2$AM\cite{sofiiuk2021foreground}, DHT+\cite{guo2022transformer}) on RealHM. We see from the first five rows that our method can better harmonize a composite image than other methods. We show a controversial example in the last row. Different people may have different opinions regarding which harmonized image looks more natural.}
    \vspace{-2pt}
    \label{fig:examples3}
\end{figure*}

\noindent In Figure 7 of the main paper, we show that only using shifted window (Swin) attention \textit{occasionally} causes block-shaped artifacts. We explain the reason why the block-shaped artifacts appear. In Figure \ref{fig:block}, we present an illustration of the Swin attention. Visual tokens within the same window (shown in the same color) can attend to each other, but cannot attend to their neighboring tokens in other windows. For example, the two tokens that contain circles cannot attend to each other, even if they are next to each other. This may cause the block-shaped artifacts, and motivates our proposed use of global attention to address this challenge.

\section{Additional Ablation Studies}

\begin{table}[h]
	\centering
	\begin{tabular}{c|c||ccc}
        \specialrule{.1em}{.05em}{.05em} 
        \rowcolor{mygray} & & \multicolumn{3}{c}{pre-training dataset} \\
		\rowcolor{mygray} dataset & metric & COCO & OI & COCO ($50\%$) \\
        \hline\hline
		\multirow{4}[2]{*}{all} & PSNR$\uparrow$ & 39.0 & 38.9 & 38.5 \\
		& MSE$\downarrow$ & 20.9 & 21.5 & 23.7 \\
		& fPSNR$\uparrow$ & 26.6 & 26.5 & 26.1 \\
		& fMSE$\downarrow$ & 250.0 & 253.7 & 280.0 \\
        \specialrule{.1em}{.05em}{.05em} 
	\end{tabular}
    \vspace{-2pt}
	\caption{Comparison of the performance of our method pre-trained using three different datasets, i.e., COCO, Open Images V6 (denoted as OI) and $50\%$ of COCO dataset.}
	\vspace{-8pt}
	\label{tab:dataset}%
\end{table}%

\noindent We investigate the influence of the pre-training dataset to the performance of our method. We present a comparison of the performance of our method pre-trained using three different datasets, \ie, the unlabeled set from MS COCO (denoted as COCO), 120K images from Open Images V6 (denoted as OI) and $50\%$ of the unlabeled set from MS COCO (denoted as $50\%$ COCO) in Table \ref{tab:dataset}.

We see that pre-training SwinIH with our method (LEMaRT) on a set of 120K randomly sampled images from Open Images V6 (roughly of same size as the unlabeled set of MS COCO). The results are comparable to those of SwinIH pre-trained on MS COCO (only $0.1$ dB lower PSNR and $0.7$ higher MSE on iHarmony4). This indicates that common images from the Internet can be used for LEMaRT pre-training. We see that training our model on $50\%$ of COCO results in 0.4 dB drop in PSNR relative to pre-training on $100\%$ of COCO. This shows the benefit of using a larger pre-training dataset.

\section{Training Time and GPU Requirement}

\noindent Pre-training our model (SwinIH) does not require additional hardware and pre-training time scales linearly with dataset size. For example, pre-training SwinIH on COCO takes $54\%$ of the time required to train/fine-tune SwinIH on iHarmony4 dataset. Compared to another Transformer-based harmonization model DHT+ \cite{guo2022transformer}, SwinIH uses $18\%$ less time and $12\%$ less GPU memory, underscoring the efficiency of our architecture.

\section{User Study}

\noindent We conducted a limited user study of our model compared to DHT+ \cite{guo2022transformer} using real data to complement our quantitative results in the main paper. We randomly sampled $50$ real composite images from RealHM. Using the method in \cite{Hang2022SCSCo}, we had $7$ participants who rated $1050$ image pairs. The normalized B-T score for LEMaRT was $51.31$, and $48.7$ for DHT+, indicating our better qualitative performance. In our future work, we will conduct a more comprehensive user study with more participants and a larger number of images.

\section{Acknowledgement}

\noindent We sincerely thank Shixing Chen, Sheik Dawood, Chun-Hao Liu, Sajal Maheshwari and Efram Potelle for their help with the user study.

\setlength{\tabcolsep}{3pt}
\renewcommand{\arraystretch}{1.1}
\begin{table*}[t]
	\centering
	\begin{tabular}{c|c|ccccccccccc}
        \specialrule{.1em}{.05em}{.05em} 
		\rowcolor{mygray}  &  & composite & DIH & S$^2$AM & DoveNet & BargNet & IntrHarm  & RainNet & iS$^2$AM & DHT+ & SCS-Co & LEMaRT \\
		\rowcolor{mygray} dataset & metric & image &~\cite{tsai2017deep} &~\cite{cun2020improving} &~\cite{Cong2020DoveNet} &~\cite{cong2021bargainnet} &~\cite{Guo2021Intrinsic} &~\cite{Ling2021RAINNet} &~\cite{sofiiuk2021foreground} &~\cite{guo2022transformer} &~\cite{Hang2022SCSCo} & [SwinIH] \\
        \hline \hline
		\multirow{4}[2]{*}{HCOCO} & \textcolor{mygray2}{PSNR$\uparrow$}  & \textcolor{mygray2}{33.9} & \textcolor{mygray2}{34.7} & \textcolor{mygray2}{35.5} & \textcolor{mygray2}{35.8} & \textcolor{mygray2}{37.0} & \textcolor{mygray2}{37.2} & \textcolor{mygray2}{37.1} & \textcolor{mygray2}{39.2} & \textcolor{mygray2}{39.2} & \textit{\textcolor{mygray2}{39.9}} & \better{\textbf{\textcolor{mygray2}{41.0}}}{1.1}  \\
		& \textcolor{mygray2}{MSE$\downarrow$}   & \textcolor{mygray2}{69.4} & \textcolor{mygray2}{51.9} & \textcolor{mygray2}{41.1} & \textcolor{mygray2}{36.7} & \textcolor{mygray2}{24.8} & \textcolor{mygray2}{24.9} & \textcolor{mygray2}{29.5} & \textcolor{mygray2}{{16.5}} & \textcolor{mygray2}{15.0} & \textit{\textcolor{mygray2}{13.6}} & \betterdown{\textbf{\textcolor{mygray2}{10.1}}}{3.5} \\
		& fPSNR$\uparrow$ & 19.9 & 20.7 & 22.5 & 22.5 & - & 24.0 & 22.4 & - & 25.8 & - & \better{\textbf{26.9}}{1.1} \\
		& fMSE$\downarrow$  & 996.6 & 799.0 & 542.1 & 551.0 & 397.9 & 416.4 & 501.2 & {266.2} & 274.6 & \textit{245.5} & \betterdown{\textbf{209.4}}{36.1} \\
        \hline\hline
		\multirow{4}[2]{*}{HAdobe} & \textcolor{mygray2}{PSNR$\uparrow$}  & \textcolor{mygray2}{8.2} & \textcolor{mygray2}{32.3} & \textcolor{mygray2}{33.8} & \textcolor{mygray2}{34.3} & \textcolor{mygray2}{35.3} & \textcolor{mygray2}{35.2 } & \textcolor{mygray2}{36.2} & \textcolor{mygray2}{38.1} & \textcolor{mygray2}{37.2} & \textit{\textcolor{mygray2}{38.3}} & \better{\textbf{\textcolor{mygray2}{39.4}}}{1.1} \\
		& MSE$\downarrow$   & \textcolor{mygray2}{345.5} & \textcolor{mygray2}{92.7} & \textcolor{mygray2}{63.4} & \textcolor{mygray2}{52.3} & \textcolor{mygray2}{39.9} & \textcolor{mygray2}{43.0} & \textcolor{mygray2}{43.4} & \textcolor{mygray2}{21.9} & 3\textcolor{mygray2}{6.8} & \textit{\textcolor{mygray2}{21.0}} & \betterdown{\textbf{\textcolor{mygray2}{18.8}}}{2.2} \\
		& fPSNR$\downarrow$ & 17.5 & 22.4 & 24.3 & 25.1 & - & 25.9 & 25.0 & - & \textit{27.1} & - & \better{\textbf{29.2}}{2.1} \\
		& fMSE$\downarrow$  & 2051.6 & 593.0 & 404.6 & 380.4 & 279.7 & 284.2 & 317.6 & {174.0} & 242.6 & \textit{165.5} & \betterdown{\textbf{147.3}}{18.2} \\
        \hline\hline
		\multirow{4}[2]{*}{HFlickr} & \textcolor{mygray2}{PSNR$\uparrow$}  & \textcolor{mygray2}{28.3} & \textcolor{mygray2}{29.6} & \textcolor{mygray2}{30.0} & \textcolor{mygray2}{30.2} & \textcolor{mygray2}{31.3} & \textcolor{mygray2}{31.3} & \textcolor{mygray2}{31.6} & \textcolor{mygray2}{33.6} & \textcolor{mygray2}{33.6} & \textit{\textcolor{mygray2}{34.2}} & \better{\textbf{\textcolor{mygray2}{35.3}}}{1.1} \\
		& \textcolor{mygray2}{MSE$\downarrow$}   & \textcolor{mygray2}{264.4} & \textcolor{mygray2}{163.4} & \textcolor{mygray2}{143.5} & \textcolor{mygray2}{133.1} & \textcolor{mygray2}{97.3} & \textcolor{mygray2}{105.1} & \textcolor{mygray2}{110.6} & \textcolor{mygray2}{69.7} & \textcolor{mygray2}{67.9} & \textit{\textcolor{mygray2}{55.8}} & \betterdown{\textbf{\textcolor{mygray2}{40.7}}}{15.1} \\
		& fPSNR$\uparrow$ & 18.1 & 19.3 & 20.9 & 20.8 & - & 21.6 & 21.0 & - & 23.5 & - & \better{\textbf{25.0}}{1.5} \\
		& fMSE$\downarrow$  & 1574.4 & 1099.1 & 785.7 & 827.0 & 698.4 & 716.6 & 688.4 & {443.7} & 471.1 & \textit{393.7} & \betterdown{\textbf{342.7}}{51.0} \\
        \hline\hline
		\multirow{4}[2]{*}{HD2N} & \textcolor{mygray2}{PSNR$\uparrow$}  & \textcolor{mygray2}{34.0} & \textcolor{mygray2}{34.6} & \textcolor{mygray2}{34.5}  & \textcolor{mygray2}{35.3} & \textcolor{mygray2}{35.7} & \textcolor{mygray2}{36.0} & \textcolor{mygray2}{34.8} & \textcolor{mygray2}{37.7} & \textcolor{mygray2}{36.4} & \textit{\textcolor{mygray2}{37.8}} & \better{\textbf{\textcolor{mygray2}{38.1}}}{0.3} \\
		& \textcolor{mygray2}{MSE$\downarrow$}   & \textcolor{mygray2}{109.7} & \textcolor{mygray2}{82.3} & \textcolor{mygray2}{76.6} & \textcolor{mygray2}{52.0} & \textcolor{mygray2}{51.0} & \textcolor{mygray2}{55.5} & \textcolor{mygray2}{57.4} & \textbf{\textcolor{mygray2}{40.6}} & \textcolor{mygray2}{49.7} & \textit{\textcolor{mygray2}{41.8}} & \worseup{\textcolor{mygray2}{42.3}}{1.7} \\
		& fPSNR$\uparrow$ & 19.1 & 19.7 & 20.5 & 20.6 & - & 21.7 & 20.2 & - & 21.7 & - & \better{\textbf{22.8}}{1.1} \\
		& fMSE$\downarrow$  & 1410.0 & 1129.4 & 989.1 & 1075.7 & 835.6 & 797.0 & 916.5 & \textit{591.0} & 736.6 & {606.8} & \betterdown{\textbf{580.5}}{10.5} \\
        \hline\hline
		\multirow{4}[2]{*}{all} & \textcolor{mygray2}{PSNR$\uparrow$}  & \textcolor{mygray2}{31.6} & \textcolor{mygray2}{33.4} & \textcolor{mygray2}{34.3} & \textcolor{mygray2}{34.8} & \textcolor{mygray2}{35.9} & \textcolor{mygray2}{35.9} & \textcolor{mygray2}{36.1} & \textcolor{mygray2}{38.2} & \textcolor{mygray2}{37.9} & \textit{\textcolor{mygray2}{38.8}} & \better{\textbf{\textcolor{mygray2}{39.8}}}{1.0} \\
		& \textcolor{mygray2}{MSE$\downarrow$}   & \textcolor{mygray2}{172.5} & \textcolor{mygray2}{76.8} & \textcolor{mygray2}{59.7} & \textcolor{mygray2}{52.3} & \textcolor{mygray2}{37.8} & \textcolor{mygray2}{38.7} & \textcolor{mygray2}{40.3} & \textcolor{mygray2}{24.4} & \textcolor{mygray2}{27.9} & \textit{\textcolor{mygray2}{21.3}} & \betterdown{\textbf{\textcolor{mygray2}{16.8}}}{4.5} \\
		& fPSNR$\uparrow$ & 19.0 & 21.0 & 22.8 & 23.0 & - & 24.2 & 23.0 & - & \textit{25.9} & - & \better{\textbf{27.2}}{1.3}\\
		& fMSE$\downarrow$  & 1376.4 & 773.2 & 594.7 & 532.6 & 405.2 & 400.3 & 469.6 & {265.0} & 295.6 & \textit{248.9} & \betterdown{\textbf{213.3}}{35.6} \\
        \specialrule{.1em}{.05em}{.05em} 
	\end{tabular}
    \vspace{-2pt}
	\caption[]{Our image harmonization method, LEMaRT [SwinIH], outperforms state-of-the-art (SOTA) methods on iHarmony4 across \textbf{all} four metrics including fPSNR and fMSE (due to space constraints, fPSNR and fMSE are omitted in Table 1 of the main paper). PSNR and MSE are shown in {\textcolor{mygray2}{gray}}, as they have been shown in the Table 1 of the main paper. We repeat them for the readers' convenience. The column named \textit{composite image} shows the results for the direct copy and paste of foreground regions on top of background images.}
	\label{tab:sota-ihd}%
\end{table*}%

\setlength{\tabcolsep}{8pt}
\renewcommand{\arraystretch}{1.1}
\begin{table*}[t]
	\centering
	\begin{tabular}{c|c|ccccccccccc}
        \specialrule{.1em}{.05em}{.05em} 
		\rowcolor{mygray}  &  & FRIH & CDTNet & S$^{2}$CRNet & Harmonizer & LEMaRT \\
		\rowcolor{mygray} dataset & metric &~\cite{peng2022frih} &~\cite{Cong2022HighResolutionIH}&~\cite{Liang2022SpatialSeparated} &~\cite{Ke2022Harmonizer} & [SwinIH] \\
        \hline \hline
		\multirow{2}[2]{*}{HCOCO} & PSNR$\uparrow$  & 39.4 & 39.2 & 38.5 & 38.8 & \better{\textbf{41.0}}{1.6}  \\
		& MSE$\downarrow$   & 15.1 & 16.3 & 23.2 & 17.3 & \betterdown{\textbf{10.1}}{5.0} \\
        \hline\hline
		\multirow{2}[2]{*}{HAdobe} & PSNR$\uparrow$  & 37.7 & 38.2 & 36.4 & 37.6 & \better{\textbf{39.4}}{1.2} \\
		& MSE$\downarrow$   & 23.6 &  20.6 & 34.9 & 21.9 & \betterdown{\textbf{18.8}}{1.8} \\
        \hline\hline
		\multirow{2}[2]{*}{HFlickr} & PSNR$\uparrow$  & 33.5 & 33.6 & 32.5 & 33.6 & \better{\textbf{35.3}}{1.7} \\
		& MSE$\downarrow$   & 68.9 & 68.6 & 98.7 & 64.8 & \betterdown{\textbf{40.7}}{27.9} \\
        \hline\hline
		\multirow{2}[2]{*}{HD2N} & PSNR$\uparrow$  & 37.9 & 38.0 & 36.8 & 37.6 & \better{\textbf{38.1}}{0.1} \\
		& MSE$\downarrow$   & 42.8 & 36.7 & 51.7 & \textbf{33.1} & \worseup{42.3}{8.8} \\
        \hline\hline
		\multirow{2}[2]{*}{all} & PSNR  & 38.2 & 38.2 & 37.2 & 37.8 & \better{\textbf{39.8}}{1.6} \\
		& MSE$\downarrow$   & 24.0 & 24.7 & 35.6 & 24.3 & \betterdown{\textbf{16.8}}{7.2} \\
        \specialrule{.1em}{.05em}{.05em} 
	\end{tabular}
    \vspace{-2pt}
	\caption[]{Comparison between our LEMaRT [SwinIH] model and four additional image harmonization models~\cite{peng2022frih},~\cite{Cong2022HighResolutionIH},~\cite{Liang2022SpatialSeparated},~\cite{Ke2022Harmonizer} on iHarmony4. We include these models for completeness, although they underperform the state-of-the-art (SOTA) method~\cite{Hang2022SCSCo} presented in the main paper.}
	\label{tab:sota-eccv}%
\end{table*}%

{\small
\clearpage
\bibliographystyle{ieee_fullname}
\bibliography{references}
}